\documentclass[letterpaper, 10 pt, conference]{ieeeconf}  % Comment this line out if you need a4paper

\IEEEoverridecommandlockouts                              % This command is only needed if 
\usepackage{multicol}
\usepackage[bookmarks=true]{hyperref}
\usepackage{stix}
\usepackage{amsmath} % assumes amsmath package installed
\usepackage{amssymb}  % assumes amsmath package installed
\usepackage{xcolor}
\usepackage{xspace}
\usepackage{filecontents}

\definecolor{ceil}{rgb}{0.16, 0.32, 0.75}
\definecolor{ceruleanblue}{rgb}{0.16, 0.32, 0.75}
\definecolor{chestnut}{rgb}{0.8, 0.36, 0.36}

\newcommand{\note}[1]{{\footnotesize \textcolor{ceil}{$\triangleright$ #1}}}

\newcommand{\sz}[1]{\textcolor{red}{\bf\small }}
\newcommand{\sarvesh}[1]{\textcolor{green}{\bf\small }}
\newcommand{\apdx}[1]{Appendix #1}

\usepackage[normalem]{ulem}

\usepackage{verbatim}
\usepackage{algorithm}
\usepackage{algorithmicx}

\usepackage{algpseudocode}
\usepackage{graphicx}
\usepackage{adjustbox}
\usepackage{tikz}
\usepackage{pgfplots}
\usepackage{textcomp}
\usepackage{xcolor}
\usepackage{booktabs}
\usepackage{siunitx}

\usepackage{multirow,array}

\usepackage{subcaption}
\usepackage{hyperref}
\usepackage{footmisc}
\usepackage{subfloat}

\makeatletter
\newcommand\fs@spaceruled{\def\@fs@cfont{\bfseries}\let\@fs@capt\floatc@ruled
  \def\@fs@pre{\vspace{5\baselineskip}\hrule height.8pt depth0pt \kern2pt}%
  \def\@fs@post{\kern2pt\hrule\relax}%
  \def\@fs@mid{\kern2pt\hrule\kern2pt}%
  \let\@fs@iftopcapt\iftrue}
\makeatother

\usepackage[utf8]{inputenc}
\usepackage[english]{babel}
\usepackage{epstopdf}
\title{ \bf
Enhancing Dexterity in Robotic Manipulation \\via Hierarchical Contact Exploration
}

\author{Xianyi Cheng, Sarvesh Patil, Zeynep Temel, Oliver Kroemer, and Matthew T. Mason% <-this % stops a space \\
\thanks{The authors are with Carnegie Mellon University, Pittsburgh, PA, 15213, USA. 
{\tt\small \{xianyic, sarveshp, ztemel, okroemer, mm3x\}@andrew.cmu.edu}%
\newline
Video \url{https://youtu.be/fScfat1Ys6U} \newline
Code \url{https://github.com/XianyiCheng/HiDex} \newline
Website \url{https://xianyicheng.github.io/HiDex-Website}}
}

\begin{document}
\maketitle

\begin{abstract}
Planning robot dexterity is challenging due to the non-smoothness introduced by contacts, intricate fine motions, and ever-changing scenarios.
We present a hierarchical planning framework for dexterous robotic manipulation (HiDex). This framework explores in-hand and extrinsic dexterity by leveraging contacts. It generates rigid-body motions and complex contact sequences. 
Our framework is based on Monte-Carlo Tree Search (MCTS) and has three levels: 
1) planning object motions and environment contact modes; 
2) planning robot contacts; 
3) path evaluation and control optimization. 
This framework offers two main advantages. 
First, it allows efficient global reasoning over high-dimensional complex space created by contacts. 
It solves a diverse set of manipulation tasks that require dexterity, both intrinsic (using the fingers) and extrinsic (also using the environment), mostly in seconds.  
Second, our framework allows the incorporation of expert knowledge and customizable setups in task mechanics and models. It requires minor modifications to accommodate different scenarios and robots. Hence, it provides a flexible and generalizable solution for various manipulation tasks. 
As examples, we analyze the results on \emph{7} hand configurations and \emph{15} scenarios. We demonstrate \emph{8} tasks on two robot platforms. 
\end{abstract}
\section{Introduction}\label{sec:intro}

Robots need dexterity for complex manipulation tasks. 
Consider taking a book from the bookshelf. The robot should consider the occlusion of the bookshelf and other books, even use them, to get the book out. The robot needs to not only use its own fingers dexterously, but also be smart about exploiting its surroundings, as ``external'' fingers to support the movements of the object. 

Planning for dexterity remains challenging. First, planning through contacts, which involves changes in system dynamics and non-smoothness, is particularly difficult \cite{posa2014direct}, especially considering both robot and environment contacts. Second, due to the diverse nature of manipulation, robots need to discover various fine motions, making it hard to simplify problems with predefined primitives. Third, current manipulation planners are tailored for specific tasks. An in-hand manipulation policy \cite{andrychowicz2020learning} cannot solve object reorientation in \cite{hou2018fast}, and neither of them can be directly applied to planar pushing \cite{lynch1996stable}. As real-world manipulation is a mix of various manipulation problems, it is important for a general manipulation planner to cover different tasks.  

\begin{figure}[!t]
    \centering
    \includegraphics[width = 0.95\columnwidth]{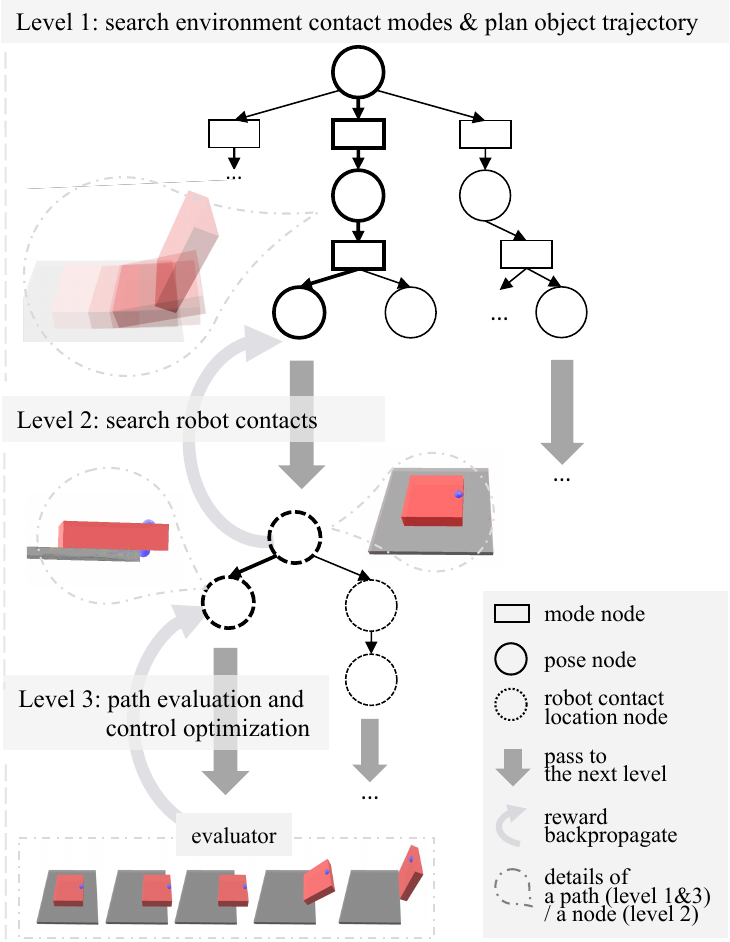}
    \caption{An overview of our framework, with an example of picking up a card. The following processes run iteratively. Level 1 plans object trajectories, interleaving searches over contact modes ($\hrectangle$ nodes) and continuous object poses ($\lgwhtcircle$ nodes). 
    An object trajectory is passed to Level 2 ($\downwhitearrow$) to plan robot contacts on the object($\dottedcircle$ nodes). The full trajectory of object motions and robot contacts is passed to Level 3 ($\downwhitearrow$) for evaluation and control optimization. After evaluation, Level 3 passes the reward back to the upper levels ($\cwopencirclearrow$). The reward is updated for every node in the path (bold nodes). In the example, the robot pulls the card to the edge of the table and then grasps it. }
    \label{fig:overview}
    \vspace{-0.5cm}
\end{figure}

We propose a \emph{hierarchical} framework, as shown in Figure~\ref{fig:overview}, aiming to address the challenges mentioned above. 
In Level 1, we perform object trajectory and environment contact mode planning. Contact modes of object-environment contacts guide the generation of object motions --- as the active exploration of extrinsic dexterity. 
In Level 2, given an object trajectory, the intrinsic dexterity is planned by optimizing for robot contact sequences on the object surface. 
In Level 3, more details of the plans are computed and rewards are backpropagated. 

Our contributions are in three aspects: framework, methodology, and experiment. 
\textbf{Framework: }
Our framework adapts with minor adjustments to various tasks. It easily encodes expert knowledge and priors through the MCTS action policies, value estimations, and rewards. This is the first framework that achieves manipulation tasks of such complexity and variety. For future development, it can directly integrate new components like trajectory optimization and learning. 
\textbf{Methodology:} 
We demonstrate the efficacy of three novel ideas with potential benefits for contact-rich robotic research: 1) an efficient hierarchical structure that decomposes the search of environment contact modes, object trajectory, and robot contacts; 2) replacing Monte-Carlo rollout with a rapidly exploring random tree (RRT) based method, leveraging its effectiveness in goal-oriented exploration without losing the Monte-Carlo spirit; 3) a novel representation that plans contact changes rather than contacts for each step, which greatly speeds up robot contact planning.
\textbf{Experiments:} 
We instantiate and demonstrate a variety of dexterous manipulation tasks, including \textit{pick up a card}, \textit{book-out-of-bookshelf}, \textit{peg-out-of-hole}, \textit{block flipping}, \textit{muti-robot planar manipulation}, and \textit{in-hand reorientation}. Some tasks were never explored in previous works, including \textit{occluded grasp}, \textit{upward peg-in-hole}, and \textit{sideway peg-in-hole}. As a contribution to the community, we open-sourced our code for these tasks. It is also easy to configure new scenarios and adjust parameters with one \href{https://github.com/XianyiCheng/HiDex/blob/main/data/template_task/setup.yaml}{\textit{setup.yaml}} file.
\section{Related Work}
\subsection{Dexterous Manipulation Planning}

In dexterous manipulation, potential contact changes introduce a high dimensional, non-convex, discrete and continuous space to plan through. 
Contact-implicit trajectory optimization \cite{posa2014direct} performs local optimization in the complex space, which can be intractable without good initialization, and currently only works for 2D problems with shape simplifications\cite{doshi2020icra,aceituno-cabezas2020rss}. 
Alternatively, contact kinematics \cite{xiao2001automatic, Mason, huang2020efficient, huang2023autogenerated} can be leveraged to generate motions between two rigid bodies~\cite{tang2008automatic}, within a robot hand~\cite{trinkle1991dexterous}, dexterous pregrasps \cite{chen2023pregrasp}, under environment contacts \cite{cheng2019suctioncups, cheng2020contact, cheng2022contact3D, liang2022learning}, and with trajectory optimization \cite{pang2022global}. 
We build upon \cite{cheng2022contact3D}, which incorporates contact modes into an RRT to guide node expansion. We introduce a more optimized search structure, addressing three key issues: no mechanism to optimize for any reward, random sampling for robot contact planning, and an entangled configuration space of the object and the robot that results in many redundant searches. 
We introduce hierarchical search to decompose the space of object motion and robot contacts \cite{lee2015hierarchical, aceituno2022hierarchical}. Unlike previous approaches using contact states, which can have the combinatoric explosion for 3D cases, our method leverages contact modes and efficient robot contact search, extending the application from 2D polyhedrons to 3D meshes.

Reinforcement learning (RL) can discover skills like in-hand manipulation \cite{andrychowicz2020learning}, dexterous grasping \cite{zhou2022learning}, and object reorientation \cite{zhou2023learning}. RL faces the same challenges from high-dimensional complex spaces, leading to sample efficiency problems. 
Our framework's advantage lies in adapting to new tasks and discovering new skills through contact exploration, without the need for training or reward shaping.

\subsection{Monte-Carlo Tree Search}
MCTS is a heuristic search algorithm that uses random sampling for efficient exploration. AlphaGo \cite{silver2016alphago} combines MCTS with deep neural networks, achieving superhuman performance in board games like Go. 
MCTS has shown potential in contact planning, including gait planning for legged robots \cite{amatucci2022mctslegged} and robot contact sequence planning given object trajectories\cite{zhu2022efficient}. 
Our work plans not only robot contacts but also object motions and interactions with environment contacts. 
MCTS offers benefits like efficient search through vast complex space, continuous improvement through learning and self-exploration, and parallelizability.

% \section{Preliminary: MCTS}
\section{Preliminaries}
\subsection{MCTS}
\label{sec:method-mcts}
Level 1 and 2 use the MCTS skeleton in Algorithm~\ref{alg:mcts}. 
A search tree $\mathcal{T} = (\mathcal{V}, \mathcal{E})$ contains a set of nodes $\mathcal{V}$ and edges $\mathcal{E}$. A node is associated with a visited state $s$. An edge is associated with a state transition $s \xrightarrow[]{a} s'$ through action $a$. 

\begin{figure}[t]
    \centering
    \vspace{0.15cm}
    \includegraphics[width = 0.98\columnwidth]{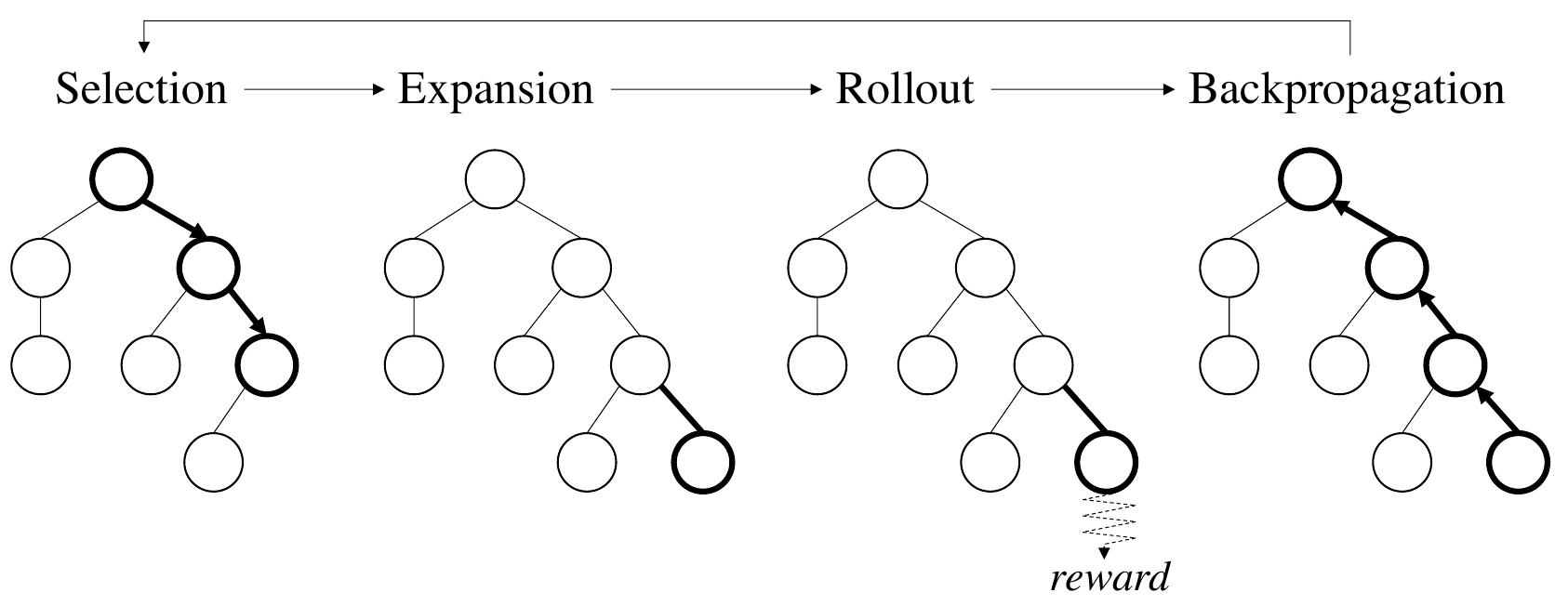}
    \caption{Four steps run iteratively in \textsc{grow-tree} in MCTS \cite{swiechowski2022monte}. Selection: start from the root node and select successive nodes. Expansion: Create a new node from unexplored children of the last selected node. Rollout: evaluate the new node by simulating to the end with random sampling. Backpropagation: Update the tree using rollout rewards.}
    \label{fig:mcts}
    \vspace{-0.3cm}
\end{figure}

\textsc{grow-tree} expands the tree by iteratively running four steps in Figure~\ref{fig:mcts}: selection, expansion, rollout, and backpropagation. 
We employ the idea in AlphaGo \cite{silver2016alphago} --- use value estimation and action probability to prioritize empirically good directions. Each node maintains a value estimation $v_{est}(s)$, obtained value $v(s)$, and the number of visits $N(s)$. 
For $s \xrightarrow[]{a} s'$, we define the action probability $p(s,a)$, the number of visits $N(s,a) = N(s')$, and the state-action value $Q(s,a) = \lambda v(s') + (1-\lambda) v_{est}(s')$, where an adaptive parameter $\lambda \in [0,1]$ balances ratios of the obtained value and the value estimation. $\lambda$ increases as the search goes on.   

Selection determines search directions by balancing exploration and exploitation. 
Among the set of feasible actions $\mathcal{A}(s)$, we select the next action $a^* \leftarrow \mathrm{argmin}_{a \in \mathcal{A}(s)} U(s,a) $ with $\eta$ controlling the degree of exploration:
\begin{equation}
\label{eqn:U}
    U(s,a) = Q(s,a) + \eta p(s,a) \frac{\sqrt{N(s)}}{1 + N(s,a)} 
\end{equation}

In backpropagation, every node on the evaluated path is updated with the reward $r$: 
\begin{equation}
\label{eqn:update_v}
    v(s) = \frac{N(s) v(s) + r}{N(s) + 1}
\end{equation}
\begin{equation}
\label{eqn:update_N}
    N(s) = N(s) + 1
\end{equation}
%A direct value estimation function is often used to calculate $v_{est}$. Otherwise if a reward estimation $r_{est}$ is used, we update $v_{est}$ with the same rule as Equation~\ref{eqn:update_v}. 
%While Level 1 and 2 share the same skeleton, we employ different designs for each level based on the nature of manipulation problems for efficient search. 

\subsection{Contact Modes}
\label{sec:cm}
Collision detection obtains the contact points between the object and the environment. 
Contact modes describe the possible evolution of these contact points. In this paper, for $N$ contacts, a contact mode is $\boldsymbol{m} = [\textit{sign}(\boldsymbol{v_{c,n}^i})] \in \{0,+\}^N$, where $\boldsymbol{v_{c,n}^i}$ is the contact normal velocity for the $i$th contact. $0$ means a contact maintained. $+$ means a contact separate.
A more comprehensive description can be found in \cite{huang2020efficient}. 

Contact modes offer an efficient way to generate object motions in lower-dimensional manifolds that have zero probability to be generated through random sampling. In addition, each contact mode corresponds to one set of continuous contact dynamics. By choosing among contact modes, discreteness in dynamics is efficiently captured. 

\begin{algorithm}[t]
\small
\caption{MCTS skeleton}\label{alg:mcts}
\begin{algorithmic}[1]
\Procedure{Search}{}
    \State $\mathcal{T} \gets$ \Call{new-tree}{}
    \State $n$ $\gets$ \Call{root-node}{$\mathcal{T}$}
    \While{$n$ is not a terminal node}
        \State \Call{grow-tree}{$\mathcal{T}, n$}
        \State $n \gets$ \Call{best-child}{$n$}
    \EndWhile
\EndProcedure
\end{algorithmic}
\end{algorithm}

\begin{figure}[t]
\vspace{-0.25cm}
\end{figure}

\section{Hierarchical Planning Framework}
\label{sec:method}

\subsection{Task Description}
\label{sec:method-task}
This paper focuses on one rigid body in a non-movable rigid environment or no environment component.  

A planner designed under this framework takes in:  
\begin{enumerate}
    \item The object: a rigid body $\mathcal{O}$ with known center of mass and inertia matrix, and friction coefficients with environment $\mu_{\mathrm{env}}$ and with the manipulator $\mu_{\mathrm{mnp}}$. The object geometry, used for collision check and surface point sampling, should be provided as a mesh model or a primitive shape like a cuboid. 
    \item Environment with known geometries, as primitive shapes or mesh models.
    \item Robot model: The robot manipulates the object through $N_{\mathrm{mnp}}$ predefined fingertip contacts. The collision models, forward and inverse kinematics, and finger contact models are known. We assume the robot makes non-sliding and non-rolling contacts. 
    \item Start specification: object start pose $\boldsymbol{x_{\mathrm{start}}} \in SE(3)$. 
    \item Goal specification: object goal region $X_{\mathrm{goal}} \subset SE(3)$.
\end{enumerate}

It outputs an object configuration trajectory $\boldsymbol{x}(t)$ and a robot control trajectory $\boldsymbol{u}(t)$.
\subsection{Level 1: Planning Environment Contact Modes and Object Trajectories}
\label{sec:method-level1}
\begin{figure*}[t]
\vspace{0.15cm}
\centering
\includegraphics[width = 0.9\textwidth]{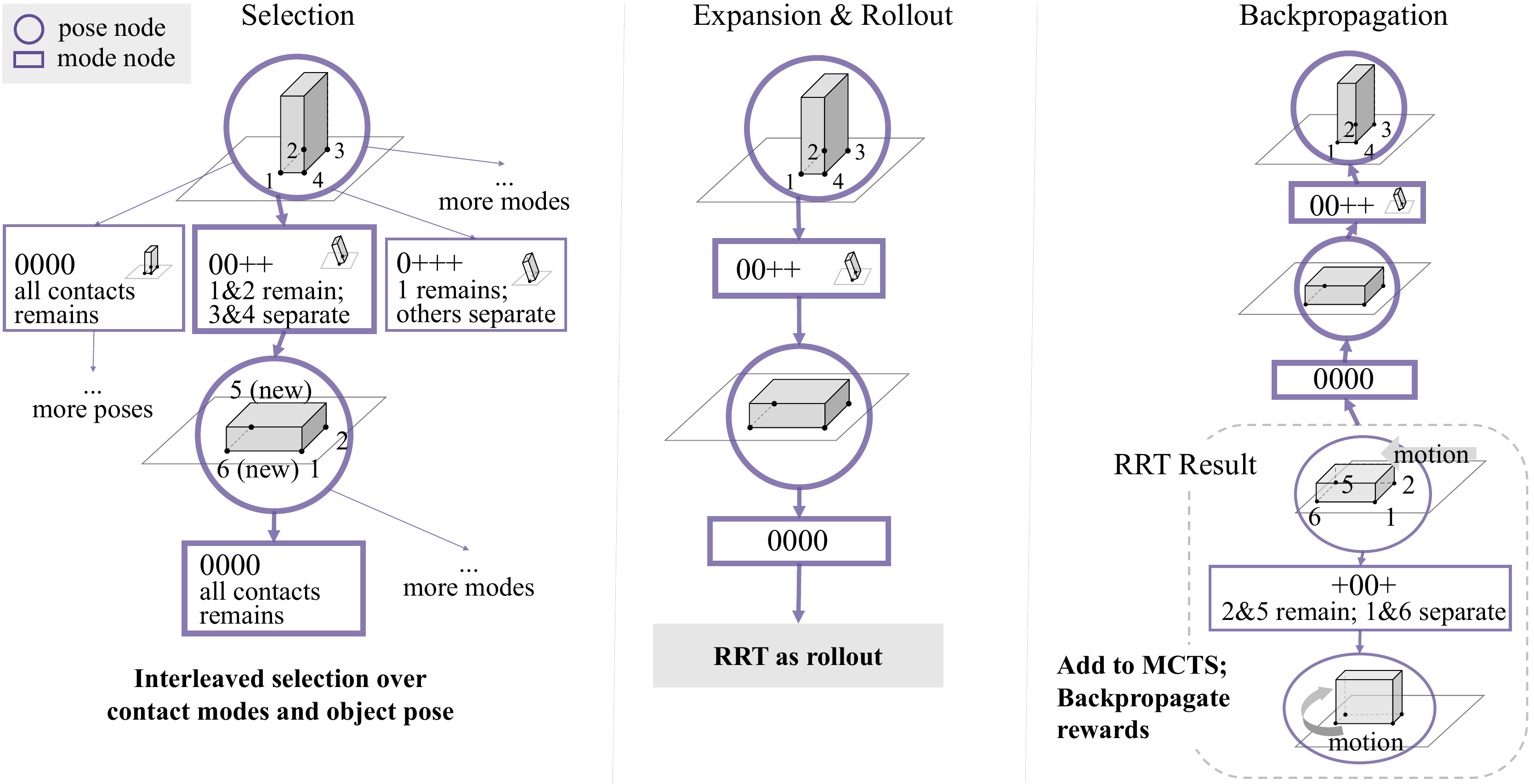}
\caption{Level 1 search visualized with a block reorientation example. 
In the selection phase, the block is selected to follow contact mode 00++, leading to a 90-degree rotation at contact 1 and 2 and the separation of 3 and 4. In the new node, contact mode 0000, indicating contact 1,2,5,6 are maintained, is then selected for the RRT rollout. The RRT generates an object configuration trajectory where the block is first pushed and then rotated 90 on the edge of contact 2 and 5. New nodes from the RRT solution path are added to the MCTS after reward evaluation.}
\label{fig:level1}
\vspace{-0.3cm}
\end{figure*}

Level 1 is summarized in Algorithm~\ref{alg:level-1}, and visualized in Figure~\ref{fig:level1}. 
It plans trajectories of environment contact modes and object configurations.
The search process is as follows. 
In the existing tree, it first performs the selection phase to choose a node to grow a new branch. We interleave the selection over environment contact modes and object configurations. When a node is selected for expansion and rollout, an RRT-based rollout replaces random rollouts to improve exploration efficiency. The RRT rollout grows a new branch for the MCTS and then passes to Level 2. 
During the process, environment contact modes are generated through a contact mode enumeration algorithm \cite{huang2020efficient}. Object configurations are generated through the projected forward integration in the RRT rollout (details in Section~\ref{sec:method-level1-rrt}).  

\begin{algorithm}[t]
    \small
    \caption{Level 1: Search Object Motion}\label{alg:level-1}
    \begin{algorithmic}[1]
    \Procedure{Grow-tree-level-1}{$\mathcal{T}, \text{startnode}$}
    \While{resources left}
        \State $n \gets $ startnode
        \State  \note{Interleaved selection over \textit{pose} and \textit{mode}}
        \While{$n$ is not terminal}
            \If{\textit{nodetype}$(n)$ is \textit{pose}}
                \State \note{[Selection] next contact mode}
                \State $\mathcal{A}(n) \gets$ feasible contact modes of $n$ 
                \State $a \gets$ \Call{select}{$\mathcal{A}(n)$}
                \State $n \gets \Call{mode-node}{a}$
            \EndIf
            \If{\textit{nodetype}$(n)$ is \textit{mode}}
                \State \note{[Selection] next pose}
                \State $\mathcal{A}(n) \gets \Call{children-of}{n} \cup \textit{explore-new}$ 
                \State $a \gets$ \Call{select}{$\mathcal{A}(n)$}
                % \State \note{[Expansion] of node $n$}
                \If{$a$ is \textit{explore-new}} \textcolor{ceil}{\footnotesize \Comment{[Expansion]}}
                    % \State \note{[Rollout] with RRT}
                    \State path $\gets$ \Call{RRT-explore}{$n$} \textcolor{ceil}{\footnotesize \Comment{[Rollout]}}
                    \State \Call{attach}{$\mathcal{T}$, path}
                    \State $r \gets $ \Call{evaluate-reward}{path} \textcolor{ceil}{\footnotesize \Comment{To Level 2}}
                    \State \Call{back-propagate}{$r, \mathcal{T}$} \textcolor{ceil}{\footnotesize \Comment{[Backpropagation]}}
                    \State break loop
                \Else
                    \State $n \gets \Call{to-next-node}{a}$
                \EndIf
            \EndIf
        \EndWhile
        \State $r \gets $ \Call{evaluate-reward}{n} \textcolor{ceil}{\footnotesize \Comment{To Level 2}}
        \State \Call{back-propagate}{$r, \mathcal{T}$} \textcolor{ceil}{\footnotesize \Comment{[Backpropagation]}}
    \EndWhile
    \EndProcedure
    \end{algorithmic}
\end{algorithm}

\subsubsection{Selection --- Interleaved Search Over Discrete and Continuous space}
\label{sec:method-level1-interleave}

We define Level 1 state as $s1 = (x\in \mathrm{SE(3)}, \textit{nodetype} \in \{ \textit{mode}, \textit{pose}\})$, where $\boldsymbol{x}$ is an object pose and \textit{nodetype} stores the type of the node. 
The interleaved selection process is demonstrated in line 6-25 in Algorithm \ref{alg:level-1} and in Figure~\ref{fig:level1}. 
For a \textit{pose} node, the action is to select a contact mode for it. The feasible actions are $\mathcal{A}\bigl(s1 = (x, \textit{pose})\bigr) = \mathcal{M}(x)$, where $\mathcal{M}(x)$ is the set of kinematically feasible contact modes enumerated for an object pose $\boldsymbol{x}$ using the algorithm in \cite{huang2020efficient}. 
A \textit{mode} node is a \textit{pose} node assigned with a contact mode. For a \textit{mode} node, the action is to select the next object pose moving from the current object pose following the contact mode. The available actions $\mathcal{A}\bigl(s1 = (x, \textit{mode})\bigr)$ comprise choosing from its child nodes (explored object poses) or \textit{explore-new}, which triggers the rollout phase to explore new object poses.

The selection policy follows Equation~\ref{eqn:U}. Action probabilities $p(s,a)$ reflect preferences of modes or poses. For example, if we prefer to exploit environment constraints to reduce uncertainties as in \cite{eppner2015exploitation}), we could have high probabilities for modes that maintain more contacts. 

\subsubsection{Expansion}
\label{sec:method-level1-expansion}
The expansion phase equals to \textit{explore-new} being selected for a \textit{mode} node. It is a variant of the progressive widening technique in MCTS for continuous space \cite{lee2020monte}, where we control the expansion rate with the action probability of \textit{explore-new}. 

\subsubsection{RRT as Rollout}

\label{sec:method-level1-rrt}

Contact-rich solutions live sparsely on lower-dimensional manifolds of the search space. It is unlikely to get any useful results from random trajectory rollouts as most traditional MCTS do. 
We replace the random rollout with an RRT search guided by contact modes (line 17, Algorithm \ref{alg:level-1}), modified based on \cite{cheng2022contact3D}. 

The RRT begins with an object pose $\boldsymbol{x_\text{current}}$ and a selected contact mode $\boldsymbol{m_\text{selected}}$. The RRT tries to reach $\boldsymbol{x_\text{goal}}$. It outputs a trajectory where each point is an object pose associated with a contact mode. 
In each iteration, we first sample an object pose $\boldsymbol{x_\text{extend}} \in \mathrm{SE(3)}$ and find its nearest neighbor $\boldsymbol{x_\text{near}}$. We then extend $\boldsymbol{x_\text{near}}$ towards $\boldsymbol{x_\text{extend}}$. Each extension is guided by a contact mode. If $\boldsymbol{x_\text{near}}$ is $\boldsymbol{x_\text{current}}$, the contact mode should be $\boldsymbol{m_\text{selected}}$. Otherwise, we enumerate all contact modes and filter them using feasibility checks. 
New object poses are generated by projected forward integration that follows each contact mode towards $\boldsymbol{x_\text{extend}}$ as close as possible. 
If the RRT finds a solution to $\boldsymbol{x_\text{goal}}$ within the maximum number of iterations, we add the solution path after the expansion node in the Level 1 search tree and proceed to Level 2 (line 19, Algorithm \ref{alg:level-1}) to obtain a reward. Otherwise, this process backpropagates a zero reward and no new node is added. 
The RRT is reused throughout the entire lifespan of the MCTS. 
More details about the RRT-based rollout can be found in \apdx{\ref*{apx:rrt}}.

One difference to \cite{cheng2022contact3D} is that we can choose whether to plan robot contacts or not in the RRT. We can turn on the option to relax a contact mode to be feasible if there ``exists any feasible robot contact'', while previously the RRT needs to plan and store robot contacts, searching in a higher-dimensional space. This relaxation could improve the planning speed for some tasks as discussed in Section~\ref{sec:results}. 

\subsection{Level 2: Planning Robot Contacts}
\label{sec:method-level2}

Level 2 initiates in \textsc{evaluate-reward} in Level 1 (line 19 and 27, Algorithm~\ref{alg:level-1}). Level 2 takes in the object configuration trajectory, and outputs the best robot contact sequence. 
Algorithm~\ref{alg:level-2} summarizes the \textsc{grow-tree} process in Level 2 MCTS. 
 
\subsubsection{State and Action Representation}
\label{sec:method-level2-representation}

Each node is associated with a robot contact state $s2 = \bigl(t, \{(i, \boldsymbol{p_i}) | i \in \text{active fingers at } t\}\bigr)$. A robot contact state specifies active fingers, fingers that are in contact with the object, and their corresponding contact locations $\{\boldsymbol{p_i} \in \mathbb{R}^3 \}$ on the object surface at timestep $t$. 
An action $a = \bigl(t_c, \{(j, \boldsymbol{p_j}) | j \in \text{relocating fingers at } t_c\}\bigr)$ represents robot contact relocations, specified by relocating timestep $t_c$, relocating fingers, and the object surface points they are relocating to $\{\boldsymbol{p_j}\}$.

For example, consider 5 available robot fingertips. Grasping an object at timestep $0$ with the first and third finger at locations $(1,0,0)$ and $(-1,0,0)$ corresponds to the state $\bigl(t=0, (\textit{finger 1}, \boldsymbol{p_1}=(1,0,0)), (\textit{finger 3}, \boldsymbol{p_3}=(-1,0,0))\bigr)$. 
If we choose to maintain the grasp until timestep 4, and then move the third finger to $(0,0,1)$ and add the fifth finger at $(-1,0,0.5)$, the action is $\bigl(t_c=4, (\textit{finger 3}, \boldsymbol{p_3}=(0,0,1)), (\textit{finger 5}, \boldsymbol{p_5}=(-1,0,0.5))\bigr)$. The new state is $\bigl(t=4, (\textit{finger 1}, \boldsymbol{p_1}=(1,0,0)), (\textit{finger 3}, \boldsymbol{p_3}=(0,0,1)), (\textit{finger 5}, \boldsymbol{p_5}=(-1,0,0.5))\bigr)$. 

Compared to the common practice of planning contacts for every timestep \cite{aceituno-cabezas2020rss, zhu2022efficient}, we plan for contact relocations. While the complexity of the search space does not change, empirically in most tasks, this modification significantly reduces the depth of the search tree and speeds up the discovery of a solution (experiments in Section~\ref{sec:results}, Figure~\ref{fig:results-level2}). 

\subsubsection{Sampling and Pruning for Action Selection}
\label{sec:method-level2-action}

Consider 100 object surface points, 4 fingers, and 10 steps. The action space is $(100^4)^{10} =1\mathrm{e}{80}$. It is not practical to evaluate all actions. To mitigate this, we adopt action sampling techniques for non-enumerable action space \cite{hubert2021learning}. 
In Equation~\ref{eqn:U}, 
we use a subset of all actions $\mathcal{A}_{\mathrm{sp}}(s2) \subset \mathcal{A}(s2)$. $\mathcal{A}_{\mathrm{sp}}(s2)$ includes previously explored actions and newly sampled actions. Newly sampled actions are generated as follows: 1) The relocating timestep $t_c$ is sampled in $(t, t_{max}]$, where $t_{max}$ is the maximum timestep the current set of contacts can proceed to under the feasibility check in Section~\ref{sec:method-level2-feasibility}. 2) After $t_c$ is sampled, we sample robot contact relocation through rejection sampling. We first find relocatable robot contacts by checking if the remaining fingers satisfy the force conditions. Then we sample new feasible contact locations on the object surface. 

We mix the selection, expansion, and rollout using the same heuristic function.
If the selected action is explored before, it is the selection phase. If the selected action is new, it is the expansion and rollout phase. 
\begin{figure}[t]
    \vspace{-0.15cm}
\end{figure}
\begin{algorithm}[t]
\small
\caption{Level 2: Search Robot Contacts}\label{alg:level-2}
\begin{algorithmic}[1]
\Procedure{Grow-tree-level-2}{$\mathcal{T}, \text{startnode}$}
\While{resources left}
    \State $n \gets $ startnode
    \While{$n$ is not terminal}
        \State $\mathcal{A}_{\mathrm{sp}}(n) \gets$ \Call{sample-feasible-actions}{n}
        \State $a \gets$ \Call{select}{$\mathcal{A}_{\mathrm{sp}}(n)$}
        \State $n \gets \Call{next-node}{n, a}$
    \EndWhile
    % \State \note{Evaluate the reward in Level 3}
    \State $r \gets $ \Call{evaluate-reward}{n} \textcolor{ceil}{\footnotesize \Comment{To Level 3}}
    \State \Call{back-propagate}{$r, \mathcal{T}$}
\EndWhile
\EndProcedure
\end{algorithmic}
\end{algorithm}
\subsubsection{Feasibility Check}
\label{sec:method-level2-feasibility}
To prune fruitless search directions, we enforce Level 2 nodes and Level 1 RRT rollout nodes to pass the feasibility check, including:
\begin{itemize}
    \item Kinematic feasibility: whether there exist inverse kinematics solutions for the robot contact points
    \item Collision-free: whether the robot links are collision-free with the environment and the object
    \item Relocation feasibility: whether there exists a plan to relocate from previous robot contacts to new contacts
    \item Force conditions: whether the chosen contact points can fulfill the task dynamics, like quasistatic, quasidynamic, force balance, or force closure conditions.
    \item Other task-specific requirements may also be added.
\end{itemize}

\subsection{Level 3: Path Evaluation and Control Optimization}
\label{sec:method-level3}
Level 3 (line 9, Algorithm \ref{alg:level-2}) computes feasibility, robot controls $\boldsymbol{u}(t)$, rewards for full trajectories from Level 2. 

If the task mechanics are quasi-static or force-closure, 
we individually solve for each step $t$ to check whether quasi-static or force-closure solutions exist and output the robot positions and optimal contact forces as the control $\boldsymbol{u}(t)$. 
If full dynamics is required, the framework can potentially use the path as a warm-start for trajectory optimization. 

For the control trajectory $\boldsymbol{u}(t)$, we compute the reward $r$ and the estimations $v_{est}$ or $r_{est}$. 
There are two rules for defining a reward function: 1) A feasible path should have a positive reward. A non-feasible path should have a zero or negative reward. It is preferred that reward values in $[0,1]$. 2) There should be a term that regularizes the length of the path. Otherwise, the search might never end.

\section{Examples and Experiments}
\label{sec:results}
\begin{figure}[t]
    \centering
    \includegraphics[width=0.9\columnwidth]{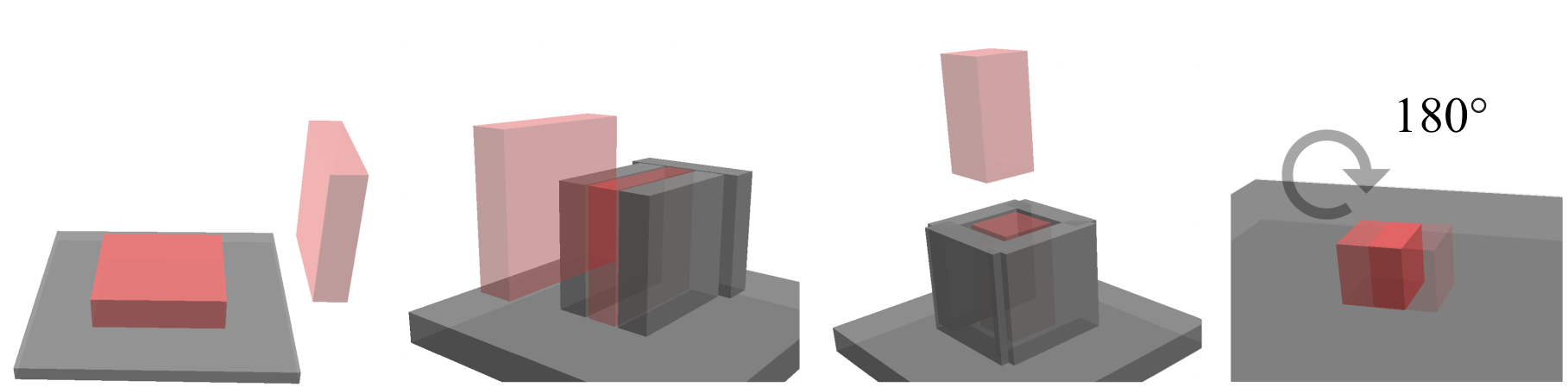}
    \caption{Manipulation with extrinsic dexterity. From left to right: 
        Scenario 1, \textit{pick up a card}: pick up a thin card that cannot be directly grasped using two available fingertips. 
        Scenario 2, \textit{bookshelf}: get a book among other books out of a bookshelf using three available fingertips. 
        Scenario 3, \textit{peg-out-of-hole}: get a peg out of a tight hole using three robot contacts (narrow gaps prevent direct grasps).  
        Scenario 4, \textit{block flipping} using two available fingertips. 
    }
    \label{fig:cmg_tasks}
    \vspace{-0.5cm}
\end{figure}

\subsection{Implementation}
We implemented two task types: manipulation with extrinsic dexterity and in-hand manipulation.
We use Dart \cite{lee2018dart} for visualization and Bullet \cite{bullet} for collision detection. 
We include new scenario setup requirements in \apdx{\ref*{apx:new}} and more implementation details in \apdx{\ref*{apx:exp}}.

\subsubsection{Robot Model}
We implemented two robot types. 

\textbf{Sphere fingertips}: Each fingertip is a sphere with workspace limits as kinematic feasibility checks. Collision is checked with the environment. 
We use three vertices of an equilateral triangle on the sphere perpendicular to the contact normal to approximate a patch contact.  

\textbf{Dexterous Direct-drive Hand} (DDHand): A DDHand has two fingers. Each fingertip has two degrees of freedom for planar translation and is equipped with a horizontal rod \cite{ddhand, bhatia2019direct}.
We provide an analytical inverse kinematics model and use the line contact model as the fingertip contact model. 

\subsubsection{Task Mechanics} 
quasi-static, quasi-dynamic, and force closure models 
(details in \apdx{\ref*{apx:exp:env:mechanics}}). 

\subsubsection{Feasibility Checks} 
include task mechanics check, finger relocation force check (during relocation, it needs to satisfy the task mechanics assuming the object is static), kinematic feasibility check, and collision check. 

\subsubsection{Features and Rewards}
We use features including travel distance ratio (total object travel distance divided by the start to goal distance), path size, robot contact change ratio (number of finger contact changes divided by the path length), and grasp centroid distance \cite{roa2015grasp}. 
Given some feature values as data points, we manually label the reward values, favoring smaller object traveling distance, less number of contact changes, and better grasp measures. Given the labeled data, we fit a logistic function as the reward function. 

\subsubsection{Action Probability}
In Level 1, in choosing the next contact mode, the action probability prioritizes the same contact mode as before: 
\begin{equation}
    p\bigl(s1 = (x, \textit{mode}),a \bigr) = 
    \begin{cases}
    0.5  \quad \text{if} \quad a = \text{previous mode}\\
    \frac{0.5}{\text{number of modes} - 1}  \quad \text{else} \quad 
    \end{cases}
\end{equation}
In choosing the next configuration, we use a uniform distribution among all the children plus the expansion action. 

In Level 2, in choosing a timestep to relocate and the contact points to relocate to, 
the action probability is calculated using a weight function $w(s2, a)$
\begin{equation}
\label{eqn: p_s_2}
    p(s2, a) = \frac{w(s2, a)}{\sum_{a' \in \mathcal{A}_{\mathrm{sp}}(s2)} w(s2, a')}
\end{equation}
$w(s2, a)$ encourages previous robot contacts to stay until $t_{\mathrm{max}}$: 
\begin{equation}
    w(s2, a) = 
    \begin{cases}
    0.5  + \frac{0.5}{t_{\mathrm{max}} - t_c + 1} \quad \text{if} \quad t_c = t_{\mathrm{max}}\\
    \frac{0.5}{t_{\mathrm{max}} - t_c + 1}  \quad \text{else} \quad 
    \end{cases}
\end{equation}

\subsubsection{Value Estimation}
We only use value estimation in Level 1. Each node has $v_{est} = 0.1$ if any Level 2 search can find a valid robot contact sequence for it. 

\subsubsection{Search Parameters}
We let $\eta = 0.1$ for both levels. We set $\lambda$ to 1 if a positive reward is found, otherwise $0$.

\begin{table}[t]
    \vspace{0.15cm}
    \centering
    \resizebox{\columnwidth}{!}{
\begin{tabular}{l|cc|cc|cc|cc}
        \toprule
          Scenario & \multicolumn{2}{c|}{1} & \multicolumn{2}{c|}{2} & \multicolumn{2}{c|}{3} & \multicolumn{2}{c}{4} \\
         \midrule
Solution found time (s) & \textbf{5.1} & 11 & \textbf{1.2} & 7.5 & \textbf{5.9} & 17 & \textbf{4.6} & 10 \\
Success rate & 1.0 & 1.0 & 1.0 & 1.0 & 1.0 & 1.0 & 1.0 & 1.0 \\
Nodes in tree & \textbf{108} & 59 & \textbf{165} & 33 & \textbf{110} & 152 & \textbf{813} & 27 \\
Solution length & \textbf{6.2} & 9.7 & \textbf{6.0} & 7.4 & \textbf{11} & 17.0 & \textbf{12} & 6.5 \\
Travel distance ratio & \textbf{1.1} & 1.4 & \textbf{1.0} & 1.2 & \textbf{1.1} & 1.8 & \textbf{1.8} & 1.2 \\
Finger relocations & \textbf{1.1} & 4.1 & \textbf{1.0} & 3.9 & \textbf{3.3} & 4.7 & \textbf{2.6} & 4.5 \\
Env contact changes & \textbf{2.4} & 2.9 & \textbf{2.0} & 2.7 & \textbf{4.6} & 4.5 & \textbf{3.2} & 4.7 \\
Grasp centroid distance & \textbf{0.2} & 0.5 & \textbf{0.9} & 1.1 & \textbf{0.6} & 0.4 & \textbf{0.3} & 0.9 \\
        \bottomrule
    \end{tabular}
    }
    \caption{Planning statistics for manipulation with extrinsic dexterity: our method (left, bold), CMGMP (right).}
    \label{tab:cmg-compare}
    \vspace{-0.5cm}
\end{table}

\subsection{Manipulation with Extrinsic Dexterity} 

We evaluate four examples in Figure~\ref{fig:cmg_tasks}. Each scenario is implemented with sphere fingertips without workspace limit and quasi-static mechanics. Additional scenarios are demonstrated in the real robot experiments in Section~\ref{sec:result-robots}. 

Table~\ref{tab:cmg-compare} shows the planning statistics from 100 runs for each scenario, using a desktop with the Intel Core i9-10900K 3.70GHz CPU (also for all other statistics in this paper). As our algorithm is anytime% (can stop anytime after a solution is found)
, we let the planner stop after 10 seconds and collect the results. The search process is fast. Similar to CMGMP\cite{cheng2022contact3D}, about 80$\%$ of the computation is on the projected forward integration of the RRT. 

\subsubsection{Ablation of Hierarchical Structure and MCTS} 

We compare with CMGMP \cite{cheng2022contact3D}, which uses an RRT in searching object motions and robot contacts. 
For all scenarios, our method has a faster ``solution found time''. The hierarchical structure speeds up solution discovery by decoupling the object pose and robot contact space. 
Our method continuously improves the solution by searching more space, whereas CMGMP can only stop upon initial solution discovery, resulting in our method generating more ``nodes in the tree''. Guided by the MCTS rewards, our method also finds solutions with a smaller ``travel distance ratio'', less ``finger relocation'' and ``environment contact changes'', and smaller ``grasp centroid distance'', while CMGMP cannot optimize for any reward.

\subsubsection{Efficient Robot Contact Planning}
\begin{figure}
\vspace{0.1cm}
\begin{tikzpicture}
\small
\begin{axis}[ybar,
        height=0.55\columnwidth,
        width=0.9\columnwidth,
        axis x line*=bottom,
        axis y line*=left,
        xlabel=Search Space Size,
        xtick={1,...,4},
        ylabel=Planning time (s),
        legend pos=north west,
        legend style={draw=none},
        xticklabels={$20^{10}$, $100^{10}$, $200^{10}$, $200^{20}$},
             nodes near coords,
             nodes near coords align={vertical},
             ymode=log,
       log origin=infty,
       log basis y={10}
             ]
\addplot[point meta=explicit symbolic, color=ceruleanblue!90!white, fill=ceruleanblue!90!white] table [x=x, y=y, meta=type, col sep=comma] {ours.csv};
\addplot[point meta=explicit symbolic, color=chestnut!90!white, fill=chestnut!90!white] table [x=x, y=y, meta=type, col sep=comma] {baseline.csv};
    \legend{Ours, w/o relocation selection}
    \end{axis}
\end{tikzpicture}
\caption{Planning time (in log scale) with respect to search space size for planning robot contacts for cube sliding. We have $\textit{search space size} = \textit{candidate contacts}^{\textit{trajectory size}}$.}\label{fig:results-level2}
\vspace{-0.5cm}
\end{figure}
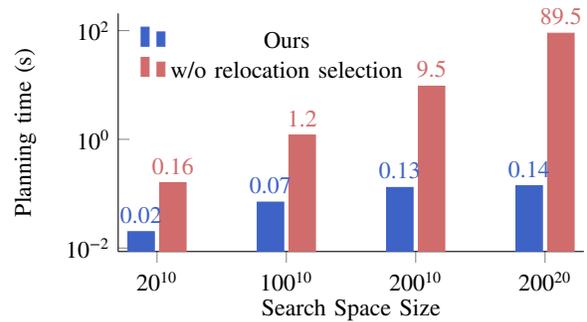

We compare the planning of contact relocations in Level 2 with the common practice that explicitly plans contacts for each timestep \cite{zhu2022efficient, aceituno-cabezas2020rss} (w/o relocation selection). 
We run the robot contact planning for a straight-line cube sliding trajectory with one allowable robot contact, with varying numbers of total timesteps in the trajectory and object surface points. 
As shown in \autoref{fig:results-level2}, the search space size grows exponentially for both methods. The planning time of the common practice also grows exponentially. Our modification uses drastically less time. Our assumption is that this modification aligns better with the fact that contact relocations are sparse compared to discretization of the entire trajectory.

\newcommand{\objectimageheight}{1cm}
\newcommand{\longobjectimageheight}{0.75cm}
\newcommand{\handimageheight}{1.25cm}
\begin{table*}[!t]
    \vspace{0.15cm}
    \resizebox{\textwidth}{!}{
    \begin{tabular}{l|ccc|cccc|cccc}
        \toprule
         Hand Type & \multicolumn{3}{c|}{3 fingers} & \multicolumn{4}{c|}{4 fingers} & \multicolumn{4}{c}{5 fingers}\\
          & \multicolumn{3}{c|}{\adjustbox{}{\includegraphics[height=\handimageheight]{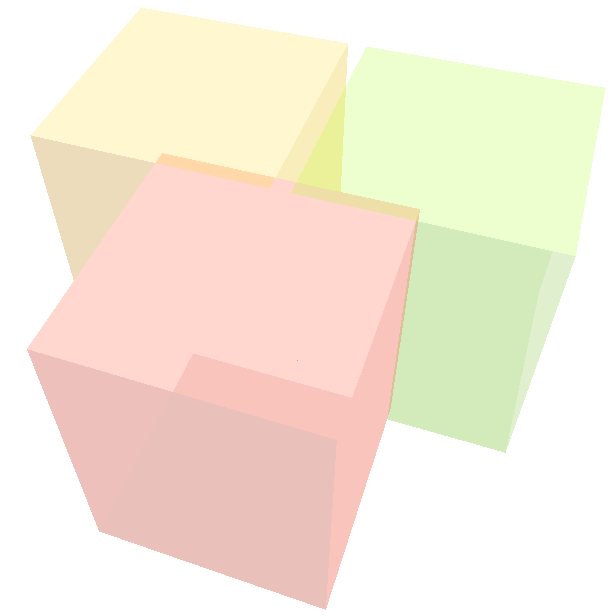}}} & \multicolumn{4}{c|}{\adjustbox{}{\includegraphics[height=\handimageheight]{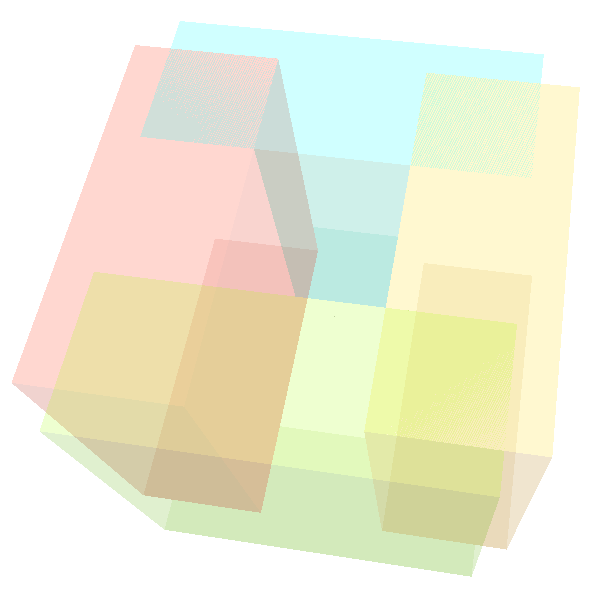}}} & \multicolumn{4}{c}{\adjustbox{}{\includegraphics[height=\handimageheight]{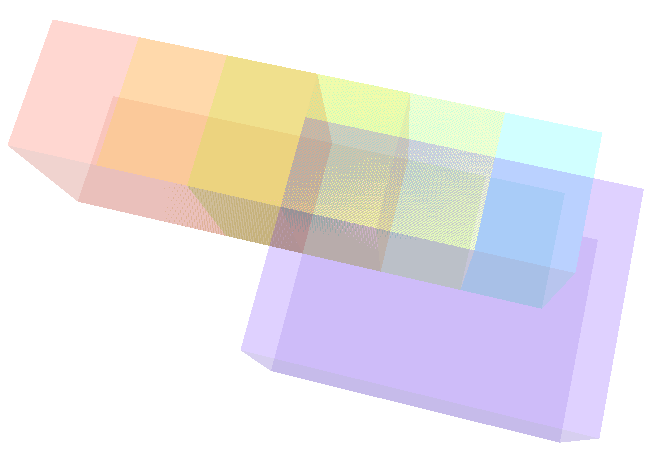}}} \\
         \midrule 
         Object & tuna fish can & b lego duplo & cylinder & apple & clamp & mug & power drill & banana & b toy airplane & hammer & c lego duplo\\
          & \adjustbox{}{\includegraphics[height=\objectimageheight]{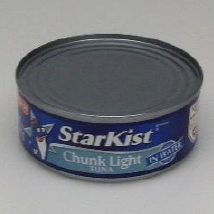}} & \adjustbox{}{\includegraphics[height=\objectimageheight]{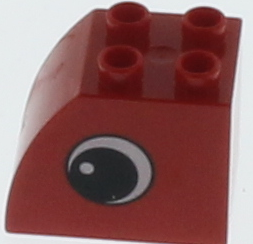}} & \adjustbox{}{\includegraphics[height=\objectimageheight]{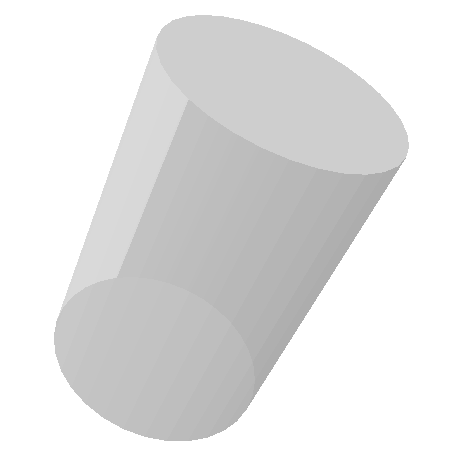}} & \adjustbox{}{\includegraphics[height=\objectimageheight]{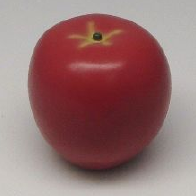}}& \adjustbox{}{\includegraphics[height=\objectimageheight]{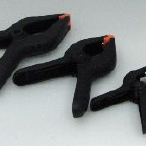}} & \adjustbox{}{\includegraphics[height=\objectimageheight]{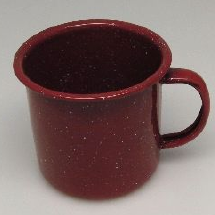}} & \adjustbox{}{\includegraphics[height=\objectimageheight]{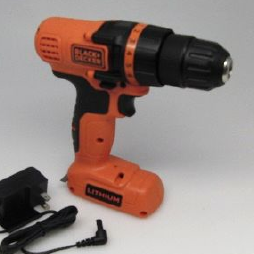}} & \adjustbox{}{\includegraphics[height=\longobjectimageheight]{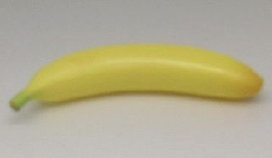}}& \adjustbox{}{\includegraphics[height=\longobjectimageheight]{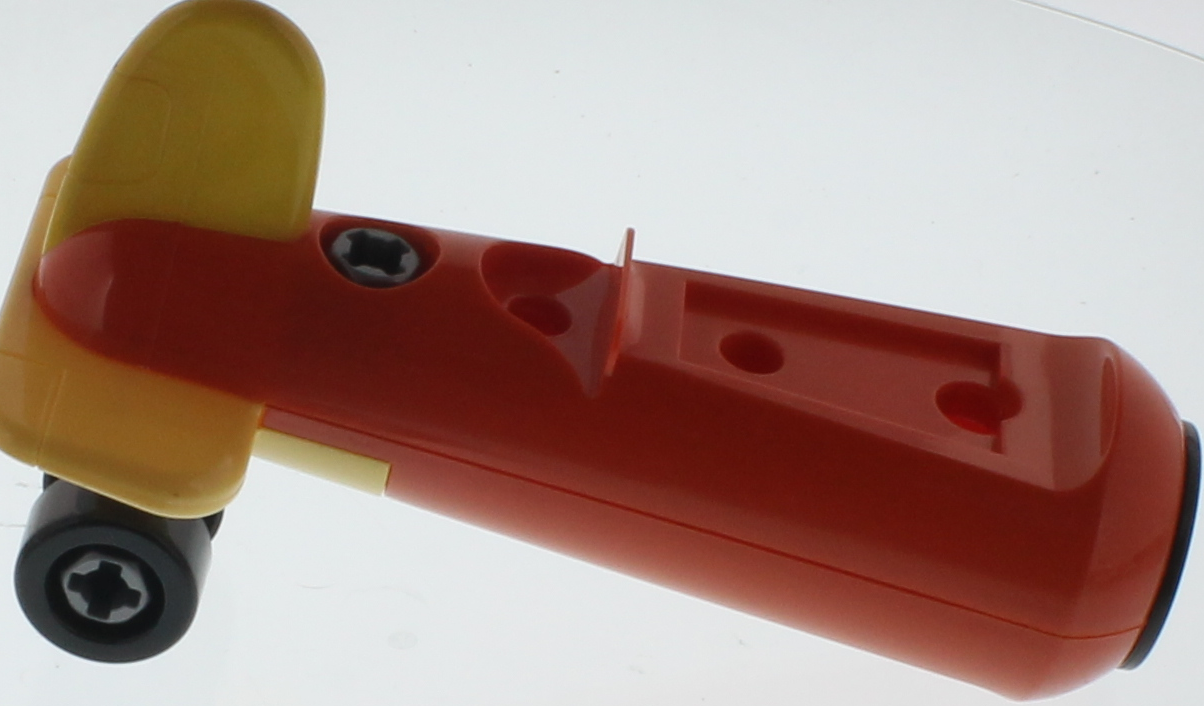}} & \adjustbox{}{\includegraphics[height=\longobjectimageheight]{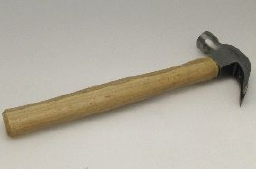}} & \adjustbox{}{\includegraphics[height=\longobjectimageheight]{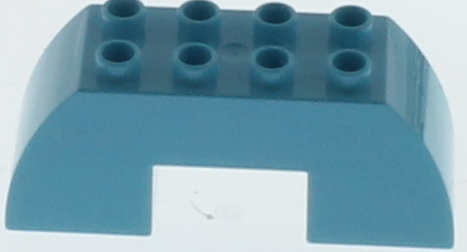}}\\
         \midrule
         Solution found time(s)  & 12.0 $\pm$ 15.3 & 4.9 $\pm$ 5.5 & 2.8 $\pm$ 4.5 & 0.3 $\pm$ 0.2 & 3.9 $\pm$ 4.9 & 0.6 $\pm$ 0.8 & 3.9 $\pm$ 4.7 & 0.6 $\pm$ 0.7 & 0.6 $\pm$ 0.6 & 0.6 $\pm$ 0.5 & 0.7 $\pm$ 1.1\\
     Success rate & 1.0 & 1.0 & 1.0 & 1.0 & 1.0 & 1.0 & 1.0 & 1.0 & 1.0 & 1.0 & 1.0\\
     Nodes in Tree & 56 $\pm$ 52 & 33 $\pm$ 15 & 50 $\pm$ 25 & 69 $\pm$ 29 & 69 $\pm$ 51 & 61 $\pm$ 29 & 72 $\pm$ 38 & 68 $\pm$ 29 & 70 $\pm$ 33 & 71 $\pm$ 30 & 63 $\pm$ 26\\
     Solution length & 10.7 $\pm$ 3.3 & 9.1 $\pm$ 2.8 & 8.2 $\pm$ 2.5 & 8.4 $\pm$ 2.1 & 9.1 $\pm$ 3.1 & 7.9 $\pm$ 2.4 & 9.7 $\pm$ 3.0 & 8.4 $\pm$ 2.3 & 8.4 $\pm$ 2.4 & 8.7 $\pm$ 2.3 & 8.4 $\pm$ 2.4\\
     Travel distance ratio & 1.5 $\pm$ 0.4 & 1.3 $\pm$ 0.3 & 1.1 $\pm$ 0.1 & 1.0 $\pm$ 0.1 & 1.1 $\pm$ 0.2 & 1.0 $\pm$ 0.1 & 1.1 $\pm$ 0.2 & 1.0 $\pm$ 0.1 & 1.0 $\pm$ 0.1 & 1.0 $\pm$ 0.1 & 1.0 $\pm$ 0.1\\
     Finger relocation & 3.1 $\pm$ 1.8 & 3.1 $\pm$ 1.9 & 3.0 $\pm$ 1.7 & 4.3 $\pm$ 1.7 & 3.4 $\pm$ 2.2 & 3.8 $\pm$ 1.7 & 3.7 $\pm$ 2.1 & 4.3 $\pm$ 1.8 & 4.2 $\pm$ 1.9 & 4.3 $\pm$ 1.8 & 4.4 $\pm$ 1.8\\
        \bottomrule
    \end{tabular}}
    \caption{Planning statistics in the format of (average $\pm$ standard deviation) for in-hand manipulation for different finger arrangements (workspaces shown by colored boxes) and objects (images from YCB dataset\cite{ycb2017}, except for the cylinder).}
    \label{tab:inhand-hands}
    \vspace{-0.5cm}
\end{table*}

\subsection{In-hand Manipulation}

In-hand manipulation demonstrates intrinsic dexterity. For inherently safer motions, we require every motion to have force balance or force closure solutions.  

\subsubsection{Different Hand Configurations} 
Table~\ref{tab:inhand-hands} shows the statistics for YCB dataset objects \cite{ycb2017} with 100 runs of randomized start and goal poses on three workspace configurations. 
Without any training or tuning, our framework achieves a high planning success rate within seconds. Point sampling on object surfaces ensures consistent performance for complex shapes, enabling planning contacts inside concave objects, like the mug and the power drill in the video. 

\subsubsection{Add Auxiliary References}

\begin{table}[t]
    \centering
       \begin{tabular}{l|cc|cc}
        \toprule
          &  \multicolumn{2}{c|}{With additional goal} & \multicolumn{2}{c}{Without}\\
          & hammer & mug & hammer & mug\\
         \midrule
            Solution found time(s)&  0.89 & 0.50 & 0.64 & 0.35 \\
            Success rate&  1.0 & 1.0 & 1.0 & 1.0 \\
            Nodes in tree&   97 & 82 & 92 & 78 \\
            Solution length&   8.7 & 8.2 & 8.5 & 8.0 \\
            travel distance ratio&   1.03 & 1.04 & 1.03 & 1.04 \\
            \textbf{Finger relocation}&   \textbf{7.4} & \textbf{5.7} & \textbf{4.9} & \textbf{4.1} \\
            \textbf{Final fingertip distance}&  \textbf{0.88} & \textbf{0.70} & \textbf{1.41} & \textbf{0.93} \\
        \bottomrule
    \end{tabular}
    \caption{Planning statistics for a 5-finger hand reorienting a hammer and a mug with and without additional goal specification of robot contact locations.}
    \label{tab:inhand-goals}
    \vspace{-0.5cm}
\end{table}

While our framework is designed to achieve object goal poses, we can incorporate auxiliary references, like goal fingertip locations, by modifying the reward function and the action probability.  
We define $d_c$, the average robot contact distance to the reference fingertip locations divided by an empirical characteristic length (like the object length). 
We fit a new reward function that prefers small $d_c$. We bias the action probability to sample contact locations that are closer to the goal through $w(s2, a)$:  
\begin{equation}
    w(s2, a) = 
    \begin{cases}
    0.5  + \frac{0.5}{t_{\mathrm{max}} - t_c + 1} p_{r}(d) \quad \text{if} \quad t_c = t_{\mathrm{max}}\\
    \frac{0.5}{t_{\mathrm{max}} - t_c + 1} p_{r}(d)  \quad \text{else} \quad 
    \end{cases}
\end{equation} 

We compare planners with and without additional goal fingertip location for 100 reorientation trials with a hammer and a mug using a 5-finger hand. Each trial has a randomized start pose, goal pose, and reference fingertip locations. As Table~\ref{tab:inhand-goals} shows, the new planner results in smaller ``Final finger distance'' to the reference fingertip locations, but more finger relocations are needed. 
Note that due to potential conflicts from the primary goal of object pose and trade-offs from other reward terms, there is no guarantee to achieve good alignment with the auxiliary goal.

\subsection{Robot Experiments}
\label{sec:result-robots}

We test 8 new scenarios on the DDHand (12kHz bandwidth) and a configurable array of soft delta robots (delta array, 500Hz control frequency) \cite{patil2022deltaarray}. We perform open-loop trajectory execution (no object pose or contact estimation). Given a planned fingertip trajectory, we compute the robot joint trajectory using inverse kinematics and execute it with joint position control. The object start position errors are within 1\si{mm} for the DDHand and 2\si{cm} for the delta array. 

\subsubsection{DDHand}

\begin{figure}[t]
    \centering
    \includegraphics[width=0.85\columnwidth]{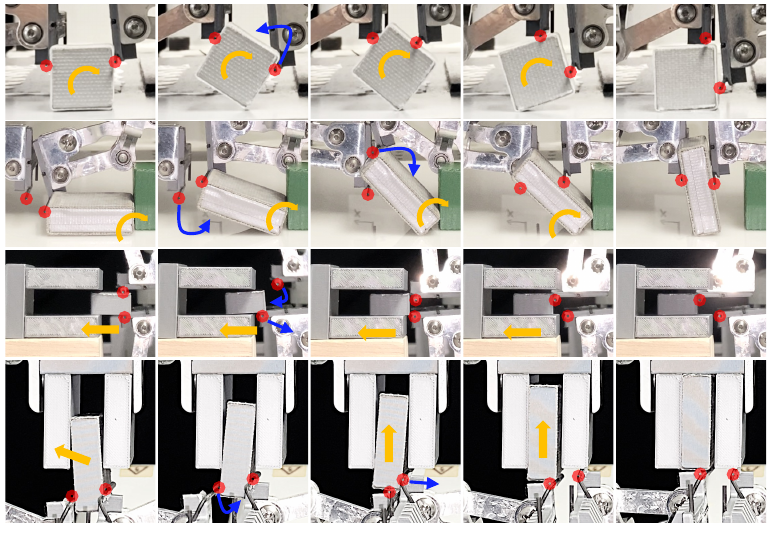}
    \caption{Keyframes of DDHand experiments. From top to bottom: \textit{cube reorientation}, \textit{occluded grasp}, \textit{sideway peg-in-hole}, \textit{upward peg-in-hole}. Object motions, fingertip locations, and fingertip relocations are respectively marked in yellow, red, and blue.}
    \label{fig:ddhand_experiments}
    \vspace{-0.3cm}
\end{figure}

The planner enables the DDHand to use intrinsic and extrinsic dexterity, as shown in Figure~\ref{fig:ddhand_experiments} and the video. 
For example, in \textit{occluded grasp}, the fixed green block and the table prevent a direct grasp. The DDHand uses three steps: pivot the object on the corner; use one finger to hold the object; and move the other finger to the other side to form a grasp. 
In \textit{upward peg-in-hole}, the hole prevents the fingers from getting in while grasping the object, but without a grasp, gravity will cause the peg to fall. 
The DDHand uses one finger to press the peg against the hole --- using the wall as an external finger to grasp. The other robot finger then relocates and pushes the peg from the bottom.

\subsubsection{Delta Array}

\begin{figure}[t]
\vspace{0.2cm}
    \centering
    \includegraphics[width=0.85\columnwidth]{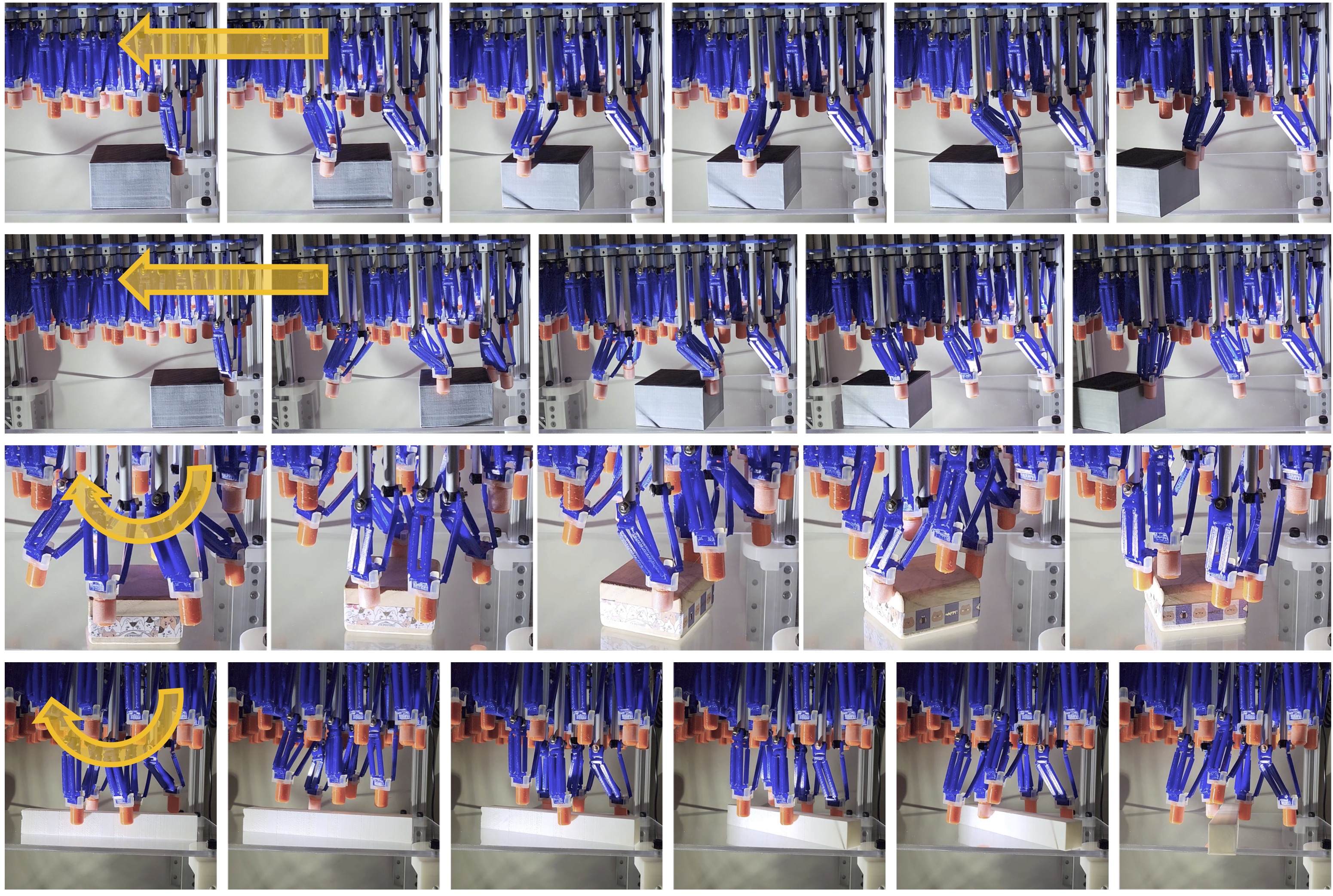}
    \caption{Keyframes of the delta array experiments. Object motions are marked with yellow arrows.  
    From top to bottom: 2-finger and 6-finger \textit{planar block passing}, 6-finger and 5-finger \textit{planar reorientation}.
    }
    \label{fig:delta_experiments}
    \vspace{-0.5cm}
\end{figure}

As shown in Figure~\ref{fig:delta_experiments} and the video, due to the small workspace of a delta robot (a cylinder with a 2 cm radius and 6cm height), many contact changes are required to accomplish the tasks. 

\section{Discussion}
\label{sec:discussion}

This paper proposes a hierarchical framework for planning dexterous robotic manipulation. 
It facilitates efficient searches across complex spaces, the generation of diverse manipulation skills, and adaptability for various scenarios. 
This method can potentially automate wide-ranging manipulation applications, such as functional grasps, caging, forceful manipulation, and mobile and aerial manipulation. For future development, this framework is compatible with direct integration of trajectory optimization, learning, complex robot contact strategies like sliding and rolling.

\bibliographystyle{IEEEtran.bst}
\bibliography{IEEEabrv, references}

\begin{thebibliography}{10}
\providecommand{\url}[1]{#1}
\csname url@samestyle\endcsname
\providecommand{\newblock}{\relax}
\providecommand{\bibinfo}[2]{#2}
\providecommand{\BIBentrySTDinterwordspacing}{\spaceskip=0pt\relax}
\providecommand{\BIBentryALTinterwordstretchfactor}{4}
\providecommand{\BIBentryALTinterwordspacing}{\spaceskip=\fontdimen2\font plus
\BIBentryALTinterwordstretchfactor\fontdimen3\font minus
  \fontdimen4\font\relax}
\providecommand{\BIBforeignlanguage}[2]{{%
\expandafter\ifx\csname l@#1\endcsname\relax
\typeout{** WARNING: IEEEtran.bst: No hyphenation pattern has been}%
\typeout{** loaded for the language `#1'. Using the pattern for}%
\typeout{** the default language instead.}%
\else
\language=\csname l@#1\endcsname
\fi
#2}}
\providecommand{\BIBdecl}{\relax}
\BIBdecl

\bibitem{posa2014direct}
M.~Posa, C.~Cantu, and R.~Tedrake, ``A direct method for trajectory
  optimization of rigid bodies through contact,'' \emph{The International
  Journal of Robotics Research}, vol.~33, no.~1, pp. 69--81, 2014.

\bibitem{andrychowicz2020learning}
O.~M. Andrychowicz, B.~Baker, M.~Chociej, R.~Jozefowicz, B.~McGrew,
  J.~Pachocki, A.~Petron, M.~Plappert, G.~Powell, A.~Ray \emph{et~al.},
  ``Learning dexterous in-hand manipulation,'' \emph{The International Journal
  of Robotics Research}, vol.~39, no.~1, pp. 3--20, 2020.

\bibitem{hou2018fast}
Y.~Hou, Z.~Jia, and M.~T. Mason, ``Fast planning for 3d any-pose-reorienting
  using pivoting,'' in \emph{2018 IEEE International Conference on Robotics and
  Automation (ICRA)}.\hskip 1em plus 0.5em minus 0.4em\relax IEEE, 2018, pp.
  1631--1638.

\bibitem{lynch1996stable}
K.~M. Lynch and M.~T. Mason, ``Stable pushing: Mechanics, controllability, and
  planning,'' \emph{The international journal of robotics research}, vol.~15,
  no.~6, pp. 533--556, 1996.

\bibitem{doshi2020icra}
F.~H. N.~Doshi and A.~Rodriguez, ``Hybrid differential dynamic programming for
  planar manipulation primitives,'' in \emph{ICRA}, 2020.

\bibitem{aceituno-cabezas2020rss}
B.~Aceituno-Cabezas and A.~Rodriguez, ``A global quasi-dynamic model for
  contact-trajectory optimization in manipulation,'' 2020.

\bibitem{xiao2001automatic}
J.~Xiao and X.~Ji, ``Automatic generation of high-level contact state space,''
  \emph{The International Journal of Robotics Research}, 2001.

\bibitem{Mason}
M.~T. Mason, \emph{Mechanics of Robotic Manipulation}.\hskip 1em plus 0.5em
  minus 0.4em\relax Cambridge, MA, USA: MIT Press, 2001.

\bibitem{huang2020efficient}
E.~Huang, X.~Cheng, and M.~T. Mason, ``Efficient contact mode enumeration in
  3d,'' in \emph{Workshop on the Algorithmic Foundations of Robotics}, 2020.

\bibitem{huang2023autogenerated}
E.~Huang, X.~Cheng, Y.~Mao, A.~Gupta, and M.~T. Mason, ``Autogenerated
  manipulation primitives,'' \emph{The International Journal of Robotics
  Research}, p. 02783649231170897, 2023.

\bibitem{tang2008automatic}
P.~Tang and J.~Xiao, ``Automatic generation of high-level contact state space
  between 3d curved objects,'' \emph{The International Journal of Robotics
  Research}, vol.~27, no.~7, pp. 832--854, 2008.

\bibitem{trinkle1991dexterous}
J.~C. {Trinkle} and J.~J. {Hunter}, ``A framework for planning dexterous
  manipulation,'' in \emph{Proceedings. 1991 IEEE International Conference on
  Robotics and Automation}, 1991, pp. 1245--1251 vol.2.

\bibitem{chen2023pregrasp}
S.~Chen, A.~Wu, and C.~K. Liu, ``Synthesize dexterous nonprehensile pregrasp
  for ungraspable objects,'' \emph{arXiv preprint arXiv:2305.04654}, 2023.

\bibitem{cheng2019suctioncups}
X.~Cheng, Y.~Hou, and M.~T. Mason, ``Manipulation with suction cups using
  external contacts,'' in \emph{Robotics Research: The 19th International
  Symposium ISRR}.\hskip 1em plus 0.5em minus 0.4em\relax Springer, 2022, pp.
  692--708.

\bibitem{cheng2020contact}
X.~Cheng, E.~Huang, Y.~Hou, and M.~T. Mason, ``Contact mode guided
  sampling-based planning for quasistatic dexterous manipulation in 2d,''
  \emph{IEEE International Conference on Robotics and Automation}, 2021.

\bibitem{cheng2022contact3D}
------, ``\href{https://ieeexplore.ieee.org/abstract/document/9811872}{Contact
  Mode Guided Motion Planning for Quasidynamic Dexterous Manipulation in 3D},''
  in \emph{2022 International Conference on Robotics and Automation (ICRA)},
  2022, pp. 2730--2736.

\bibitem{liang2022learning}
J.~Liang, X.~Cheng, and O.~Kroemer,
  ``\href{https://arxiv.org/abs/2206.12728}{Learning Preconditions of Hybrid
  Force-Velocity Controllers for Contact-Rich Manipulation},'' \emph{Conference
  on Robot Learning}, 2022.

\bibitem{pang2022global}
T.~Pang, H.~Suh, L.~Yang, and R.~Tedrake,
  ``\href{https://arxiv.org/abs/2206.10787}{Global Planning for Contact-Rich
  Manipulation via Local Smoothing of Quasi-dynamic Contact Models},''
  \emph{arXiv preprint arXiv:2206.10787}, 2022.

\bibitem{lee2015hierarchical}
G.~Lee, T.~Lozano-P{\'e}rez, and L.~P. Kaelbling, ``Hierarchical planning for
  multi-contact non-prehensile manipulation,'' in \emph{2015 IEEE/RSJ
  International Conference on Intelligent Robots and Systems (IROS)}.

\bibitem{aceituno2022hierarchical}
B.~Aceituno and A.~Rodriguez,
  ``\href{https://ieeexplore.ieee.org/document/9981862}{A Hierarchical
  Framework for Long Horizon Planning of Object-Contact Trajectories},'' in
  \emph{2022 IEEE/RSJ International Conference on Intelligent Robots and
  Systems (IROS)}.\hskip 1em plus 0.5em minus 0.4em\relax IEEE, 2022, pp.
  189--196.

\bibitem{zhou2022learning}
W.~Zhou and D.~Held, ``Learning to grasp the ungraspable with emergent
  extrinsic dexterity,'' in \emph{ICRA 2022 Workshop: Reinforcement Learning
  for Contact-Rich Manipulation}, 2022.

\bibitem{zhou2023learning}
W.~Zhou, B.~Jiang, F.~Yang, C.~Paxton, and D.~Held, ``Learning hybrid
  actor-critic maps for 6d non-prehensile manipulation,'' \emph{arXiv preprint
  arXiv:2305.03942}, 2023.

\bibitem{silver2016alphago}
D.~Silver, A.~Huang, C.~J. Maddison, A.~Guez, L.~Sifre, G.~Van Den~Driessche,
  J.~Schrittwieser, I.~Antonoglou, V.~Panneershelvam, M.~Lanctot \emph{et~al.},
  ``Mastering the game of go with deep neural networks and tree search,''
  \emph{nature}, 2016.

\bibitem{amatucci2022mctslegged}
L.~Amatucci, J.-H. Kim, J.~Hwangbo, and H.-W. Park, ``Monte carlo tree search
  gait planner for non-gaited legged system control,'' in \emph{2022
  International Conference on Robotics and Automation (ICRA)}, 2022.

\bibitem{zhu2022efficient}
H.~Zhu and L.~Righetti, ``\href{https://arxiv.org/abs/2206.09023}{Efficient
  Object Manipulation Planning with Monte Carlo Tree Search},'' \emph{arXiv
  preprint arXiv:2206.09023}, 2022.

\bibitem{swiechowski2022monte}
M.~{\'S}wiechowski, K.~Godlewski, B.~Sawicki, and J.~Ma{\'n}dziuk,
  ``\href{https://arxiv.org/abs/2103.04931}{Monte Carlo tree search: A review
  of recent modifications and applications},'' \emph{Artificial Intelligence
  Review}, pp. 1--66, 2022.

\bibitem{eppner2015exploitation}
C.~Eppner, R.~Deimel, J.~Alvarez-Ruiz, M.~Maertens, and O.~Brock,
  ``Exploitation of environmental constraints in human and robotic grasping,''
  \emph{The International Journal of Robotics Research}, vol.~34, no.~7, pp.
  1021--1038, 2015.

\bibitem{lee2020monte}
J.~Lee, W.~Jeon, G.-H. Kim, and K.-E. Kim, ``\href{lee2020monte}{Monte-carlo
  tree search in continuous action spaces with value gradients},'' in
  \emph{Proceedings of the AAAI conference on artificial intelligence}, 2020.

\bibitem{hubert2021learning}
T.~Hubert, J.~Schrittwieser, I.~Antonoglou, M.~Barekatain, S.~Schmitt, and
  D.~Silver,
  ``\href{https://proceedings.mlr.press/v139/hubert21a.html}{Learning and
  planning in complex action spaces},'' in \emph{International Conference on
  Machine Learning}.\hskip 1em plus 0.5em minus 0.4em\relax PMLR, 2021.

\bibitem{lee2018dart}
J.~Lee, M.~X. Grey, S.~Ha, T.~Kunz, S.~Jain, Y.~Ye, S.~S. Srinivasa,
  M.~Stilman, and C.~K. Liu, ``Dart: Dynamic animation and robotics toolkit,''
  \emph{Journal of Open Source Software}, vol.~3, no.~22, p. 500, 2018.

\bibitem{bullet}
E.~Coumans and Y.~Bai, ``Pybullet, a python module for physics simulation for
  games, robotics and machine learning,'' \url{http://pybullet.org},
  2016--2019.

\bibitem{ddhand}
A.~Gupta, Y.~Mao, A.~Bhatia, X.~Cheng, J.~King, Y.~Hou, and M.~T. Mason,
  ``\href{https://ieeexplore.ieee.org/stamp/stamp.jsp?arnumber=9981569}{Extrinsic
  Dexterous Manipulation with a Direct-drive Hand: A Case Study},'' in
  \emph{2022 IEEE/RSJ International Conference on Intelligent Robots and
  Systems (IROS)}.\hskip 1em plus 0.5em minus 0.4em\relax IEEE, 2022, pp.
  4660--4667.

\bibitem{bhatia2019direct}
A.~Bhatia, A.~M. Johnson, and M.~T. Mason, ``Direct drive hands: Force-motion
  transparency in gripper design,'' in \emph{Robotics: science and systems},
  2019.

\bibitem{roa2015grasp}
M.~A. Roa and R.~Su{\'a}rez, ``Grasp quality measures: review and
  performance,'' \emph{Autonomous robots}, vol.~38, no.~1, pp. 65--88, 2015.

\bibitem{ycb2017}
B.~Calli, A.~Singh, J.~Bruce, A.~Walsman, K.~Konolige, S.~Srinivasa, P.~Abbeel,
  and A.~M. Dollar, ``Yale-cmu-berkeley dataset for robotic manipulation
  research,'' \emph{The International Journal of Robotics Research}, vol.~36,
  no.~3, pp. 261--268, 2017.

\bibitem{patil2022deltaarray}
S.~Patil, T.~Tao, T.~Hellebrekers, O.~Kroemer, and F.~Z. Temel,
  ``\href{https://arxiv.org/abs/2206.04596}{Linear Delta Arrays for Dexterous
  Distributed Manipulation},'' \emph{arXiv preprint arXiv:2206.04596}, 2022.

\bibitem{liu1999forceclosure}
Y.-H. Liu, ``Qualitative test and force optimization of 3-d frictional
  form-closure grasps using linear programming,'' \emph{IEEE Transactions on
  Robotics and Automation}, vol.~15, no.~1, pp. 163--173, 1999.

\end{thebibliography}
\clearpage

\begin{appendices}

\section{Setting up New Scenarios}
\label{apx:new}

In this section, we provide an overview of what are required when setting up new scenarios. Please check our code and Appendix \ref{apx:exp} for the actual implementation. 

%careful for under-actuated robots and wheel robots
%contact jacobian is the rank of 3 (free to move with the object) 

\subsection{Applicability}
This framework can be considered for the tasks of manipulating a single rigid body object in a rigid environment. Environment components must be fixed and not movable. It can also be used when there is no environment component (in-hand manipulation). We need known models of the object, the environment, and the robot.

The robot used to manipulate the object needs to have known collision models, and forward and inverse kinematics. 
The only parts that can be used to manipulate the object are the defined ``fingertips'' on the robot. 
% The dimension of each finger's workspace must be $\geq 2$. 
% Note that this is different that degrees of freedom. 
% For example, the dimension of the workspace for the fingertips on a parallel jaw gripper attached to a 6-axis robot arm is 3, while the parallel jaw gripper only has 1 Dof.  

\subsection{Setup a new robot/hand}
\label{apx:new:robot}
Setting up a new robot is the most complicated part. Specifically for implementation in our C++ code, a new class need to be written to inherit a pre-defined abstract class \textsc{RobotTemplate}. The user need to fill some specific pure virtual functions that covers the following aspects. 

\subsubsection{Contact force models for fingertips} 
We use the point contact model for kinematics. However, as the force model for point contact might be too limited, we allow the use of other contact force models. 
%Given the point contact used for kinematics for a fingertip, the contact force model is approximated by several point contacts for force computation. 
The contact force models that currently exist in our implementation includes:
\begin{itemize}
    \item Point contact
    \item Patch contact: we first approximate the fingertips using spheres centered at the point contact locations. The radius of the spheres should approximate the radius of the contact patch for each fingertip. We approximate the patch contact using three point contacts at vertices of an equilateral triangle that is perpendicular to the contact normal and on the sphere.
    \item Line contact: we approximate the line contact model by two point contacts on twp ends of the line segment.
\end{itemize}

\subsubsection{Forward and inverse kinematics for fingertips}
The whole robot is required to have at lease 6 DOFs (including the arm). Because we require the robot be able to move with the object given an object trajectory in the space. 
The users need to provide the forward and inverse kinematics for the fingertips. 

Given the FK and IK model, we precompute the workspace for each fingertip. For general robot hands, we first sample joint angles to get the fingertip points in the workspace through forward kinematics, and then compute the convex hulls. While hands might differ, we estimate this process takes about seconds (with C++ implementation). 

\subsubsection{Robot collision model}
The users need to provide the collision model of the robot or the fingertips. If it is unlikely for the robot links to collide with the object or the environment, it is be okay to only provide the collision shape for the fingertips, which will make the computation much faster. 
Otherwise, the user could simply provide a robot URDF model. 

%If computation time is not a critical factor and there is no need to optimize using user-provided forward/inverse kinematics method and simplified collision models, 

\subsubsection{Contact relocation planner (optional)}
A contact relocation planner is required for checking whether a collision-free path exists for a finger to relocate to another contact location. 

\subsubsection{Contact sampling on the object surface (optional)} 
It is best that each fingertip are relatively independent on the kinematic side. If not, our random sampling of robot contacts on the object surface might have a very high rejection rate ($> 90\%$). In this case, we need the user to provide a method for the specific robot in order to more efficiently sample robot contacts on the object surface.

\subsubsection{Trajectory optimizer (optional)}
For the robots that are under-actuated (like wheeled robots), the users need to provide Level 3 a trajectory optimizer that finds feasible object states, robot states, and robot controls given Level 2 outputs as warm-start trajectories. 

\subsection{Setup a new task type}

After setting up the new robot, we need to enable the robot to do a certain type of tasks. Two major things to consider are task mechanics and task parameters for planning. 

\subsubsection{Task Mechanics} 
Task mechanics include the specific requirements and dynamical model required by the task. Do we have to fully exploit the dynamic property of the system? If yes, we need to have a good trajectory optimization algorithm for the manipulation system in Level 3 to ensure the solutions are feasible. If the task dynamics do not involve the integration of velocity and the robot is fully actuated, it is not necessary to provide a trajectory optimization method in Level 3. In these cases, the user only needs to write a function \texttt{\small{task-dynamics(object pose, object velocity, contact info, ...)}} that solves a one-step optimization problem. 
Examples include quasi-static, quasi-dynamic, closure methods, planar pushing, etc. 
% Many manipulation skills usually fall into the scope of quasi-static or quasi-dynamic models. We can also enforce force-closure or form-closure for safer manipulation. Caging or partial caging is also an interesting model to consider. If we consider a task where the object slides on a flat surface, it is worth considering planar pushing models \cite{mason1986mechanics, zhou2022learning} for better accuracy. 

\subsubsection{Design choices}
\label{apx:new:design}
A new task type requires several design choices to be made and some search parameters to be tuned. Once the choices are made, changing environments and objects in the same task type should not require more tuning. According to our experience, making designs and tuning parameters are relatively low-effort. We have found that the planner is not sensitive to specific numerical values for parameters.

The design choices include task features, action probability design for Level 1 and 2, reward design, and value estimation design for Level 1 and 2. 

% We first design the task features that are used in the action probability and reward. Basic features are path length in MCTS, number of robot contact relocations, and object travel distance. Task-dependent features can also be considered to drive toward desired behaviors. For example, the users could use different grasp measures as features for better grasp quality. The number of environment changes feature can encourage environment-constrained manipulation behaviors. 
% For generality, it is better that the features are normalized in a way that desired behaviors have similar values under different environments and objects.

Task features are used in the action probability and reward. Basic features are path length in MCTS, number of robot contact relocations, and object travel distance. Task-dependent features like grasp measures or environment contact changes can be added to encourage specific behaviors like better grasps and less environment contact switches. For generality, it is important to normalize the features by ensuring similar values for desired behaviors across different environments and objects.

There are three action probability functions we need to define: (1) select a contact mode in Level 1, (2) select a child (configuration node) for a mode node in Level 1, and (3) select (time to relocate, contacts to relocate to) in Level 2. For (1), we often encourage the use of the same contact mode as the previous one. For (2), we currently mostly use a uniform distribution. For (3), we would like to encourage relocating when the contacts are not feasible the next timestep. However, the definition of these probabilities is entirely up to the user.

To design the reward function, we use a simple approach that requires no tuning. 
Given some feature values as data points, we first manually label their reward values between 0 to 1 through human intuition. Next, we fit a logistic function to these data points as the reward function. 

Manually value estimation is very flexible. The value estimation in our method is often used to encourage the search to visit a node that has been visited but has not found any positive reward. For example, on the way to the goal pose, if an object pose is reachable through a sequence of contacts (check by Level 2), we can assign 0.1 as its value estimation. Our design principle is to give a small number to any node that is more likely to find a solution than others.

\subsubsection{Parameters}

The search parameters include MCTS exploration rates $\eta_1, \eta_2$ in Level 1 and 2. Adaptive parameter for value estimation $\lambda$. 
In all of our experiments, we let $\eta_1 = 0.1, \eta_2 = 0.1$. We let $\lambda = 0$ if a positive reward has not been found and otherwise $\lambda = 1$. While tuning the parameters may slightly improve the performance for specific tasks, we suspect that most of the time this is not a must. However, $\lambda$ might need some tuning if one day we have better value estimations, like using learned functions. 

\subsection{Setup a new environment}
If using our preset robots and tasks, the users can easily setup new environments and objects in one file called \textit{setup.yaml}.

When setting up a model for a new environment, it is usually adequate to use primitive geometries such as cuboids, cylinders, and spheres in the simulation environment. 
The users need to specify the shape parameters and the locations of the primitive shapes. 

\subsection{Setup a new object}

For a new object, the users need to provide the object mesh or specify the primitive shape.  
Surface points will be automatically uniformly sampled on the mesh. 
Each point $(\boldsymbol{p},\boldsymbol{n})$ is represented by its location ($\boldsymbol{p} \in R^3$) and its contact normal ($\boldsymbol{n} \in S^3$) in the object frame. 
It is usually sufficient to sample about 100 points. The computation is usually in milliseconds. 

The user also need to provide the object mass, object inertia, and friction coefficients for robot-object and environment-object contacts. 

For each new object and environment, the RRT parameters might need some changes, including the range of object positions, goal biased sampling probability, unit extend length, and the weight for rotation for the distance calculation. The RRT parameters does not require careful tuning, as long as they roughly reflect the task requirement. For example, the weight for rotation is good to be set to 1 if the object bounding box range is between 0.1 - 10 and rotation and translation are roughly of equal importance. If the object orientation is not important at all, the weight is good to be set to $0.01$ to $0.1$. The unit extend length should be larger if the object start and goal are very far from each other, otherwise the planner will be slow. And it should be smaller if the user expect many different maneuvers required for the task.    

Extra note: to avoid numerical issues, we usually scale the whole system such that the average length of the object bounding box is in the range of 1 - 10. 

\vspace{1cm}
\section{Experiment Details}
\label{apx:exp}

This section includes the details of the experiments in this paper. The first two are pure planning experiments. The latter two are robot experiments. 

\subsection{Manipulation with Environment Interactions}

\subsubsection{Robot model}
\label{apx:exp:env:robot}
We consider the robots as free-flying balls, meaning that we do not check for kinematic feasibility but do check for collision of the balls and the environment. 
For the contact force model, we use the patch contact model, described in \apdx{\ref*{apx:new:robot}}.  

\subsubsection{Task mechanics}
\label{apx:exp:env:mechanics}

We use quasi-static or quasi-dynamic models. 
For each timestep, we solve a convex programming problem to find if there exists a solution for contact force $\boldsymbol{\lambda_c}$ to satisfy the force conditions. The problem is formulated as follows: 
\begin{equation}
\label{eqn:optvel}
\begin{array}{ll}
     & \min\limits_{\boldsymbol{\lambda}} \| \epsilon \boldsymbol{\lambda}^T \boldsymbol{\lambda} \| \\
    \text{s.t.} & \text{quasistatic or quasidynamic condition} 
\end{array}
\end{equation}
where $\epsilon \boldsymbol{\lambda}^T \boldsymbol{\lambda}$ is a regularization term on the contact forces. 

The quasi-static condition requires the object to be under static force balance for a selected contact mode 
\begin{equation}
\label{eqn:static}
   \begin{bmatrix}G_1 \boldsymbol{h_1}, G_2 \boldsymbol{h_2}, \dots\end{bmatrix} \cdot \begin{bmatrix}\lambda_1, \lambda_2, \dots \end{bmatrix}^T + \boldsymbol{f_{\mathrm{external}}} = 0
\end{equation}
where $\begin{bmatrix}\lambda_1, \lambda_2, \dots \end{bmatrix}^T$ are the magnitudes of forces along active contact force directions $\begin{bmatrix}\boldsymbol{h_1}, \boldsymbol{h_2}, \dots \end{bmatrix}^T$ determined by contact modes. $\begin{bmatrix}G_1, G_2, \dots \end{bmatrix}^T$ are the contact grasp maps.
$\boldsymbol{f_{\mathrm{external}}}$ includes other forces on the object, such as gravity and other applied forces.

Quasidynamic assumption relaxes the requirement for objects to be in force balance, allowing short periods of dynamic motions. We assume accelerations do not integrate into significant velocities. In numerical integration, the object velocity from the previous timestep is $0$. The equations of motions become: 
\begin{equation}
\label{eqn:quasidynamic}
   M_o \dot{\boldsymbol{v^o}} = \begin{bmatrix}G_1 \boldsymbol{h_1}, G_2 \boldsymbol{h_2}, \dots\end{bmatrix} \cdot \begin{bmatrix}\lambda_1, \lambda_2, \dots \end{bmatrix}^T + \boldsymbol{f_{\mathrm{external}}}
\end{equation}
In discrete time, the object acceleration $\dot{\boldsymbol{v^o}}$ can be written as $\frac{\boldsymbol{v^o}}{h}$, where $h$ is the step size. The object velocity $\boldsymbol{v^o}$ is computed by solving the constrained velocity from the current pose to the goal pose under a contact mode. 

\subsubsection{Feasibility Checks} 
\begin{itemize}
    \item Task mechanics check: is passed if there exist a solution for \autoref{eqn:optvel}. 
    \item Finger relocation check: during relocation, the non-relocating robot contacts and environment contacts must also satisfy the task mechanics, assuming the object has zero velocity.
    \item Collision check: the spheres must not collide with the environment. 
\end{itemize}

\subsubsection{Features} 
We manually designed the features, as shown in Table \ref{tab:env-feature}.

\begin{table}[H]
    \centering
    \resizebox{\columnwidth}{!}{
       \begin{tabular}{l|c}
        \toprule
            Feature & Description \\
        \midrule
            Path size & {\footnotesize{node depth in the Level 1 tree}} \\
            Object travel distance ratio & $\frac{\text{total travel distance}}{dist(x_\text{start}, x_\text{goal})}$ \\
            Robot contact change ratio & $\frac{\text{number of finger contact changes}}{\text{number of fingers}}$ \\
            Number of environment contact changes & - \\
            % \multicolumn{2}{l}{Number of environment contact changes} \\
            Grasp centroid distance & $dist(c_{contact}, c_{geo})$ \\
        \bottomrule
    \end{tabular}
    }
    \caption{Features for Manipulation with Environment Interactions. $c_{contact}$: the centroid of all contact points; $c_{geo}$: the geometric center of the object.}
    \label{tab:env-feature}
\end{table}

\subsubsection{Action Probability}

In Level 1,
in choosing the next contact mode, we design the action probability to prioritize choosing the contact mode the same as before: 
\begin{equation}
    p\bigl(s1 = (x, \textit{mode}),a \bigr) = 
    \begin{cases}
    0.5  \quad \text{if} \quad a = \text{previous mode}\\
    \frac{0.5}{\mathrm{\# modes} - 1}  \quad \text{else} \quad 
    \end{cases}
\end{equation}

In Level 1, in choosing the next configuration, we let $p\bigl(s1 = (x, \textit{config}),a\bigr)$ be a uniform distribution for all the children and \textit{explore-new}

In Level 2, in choosing a timestep to relocate and the contact points to relocate, 
the action probability is calculated using a weight function $w(s2, a)$ designed for each action in $\mathcal{A}_{\mathrm{sp}}(s2)$:
\begin{equation}
\label{eqn: p_s_2}
    p(s2, a) = \frac{w(s2, a)}{\sum_{a' \in \mathcal{A}_{\mathrm{sp}}(s2)} w(s2, a')}
\end{equation}
The manually designed weight function $w(s2, a)$ prefers to let the previous robot contacts stay as long as possible: 
\begin{equation}
    w(s2, a) = 
    \begin{cases}
    0.5  + \frac{0.5}{t_{\mathrm{max}} - t_c + 1} \quad \text{if} \quad t_c = t_{\mathrm{max}}\\
    \frac{0.5}{t_{\mathrm{max}} - t_c + 1}  \quad \text{else} \quad 
    \end{cases}
\end{equation}

\subsubsection{Reward Design} 
We use all the features in Table \ref{tab:env-feature} and follow the logistic function fitting procedure as described in Appendix \ref{apx:new:design}.

\subsubsection{Value Estimation}
We only use value estimation for Level 1 nodes. Each node has $\boldsymbol{v_{\text{est}}} = 0.1$ if any subsequent Level 2 search is able to proceed past that node. 
For all Level 2 nodes, the value estimation is simply zero. 

\subsubsection{Search Parameters}

In both Level 1 and Level 2, we let the exploration rate $\eta_1, \eta_2 = 0.1$. 
Since we only have value estimation for Level 1, there is only one adaptive parameter $\lambda$ for Level 1 only. 
When no reward $> 0$ has been found, $\lambda = 0$. After any positive reward is observed, $\lambda = 1$.

\subsection{In-hand Manipulation}

\subsubsection{Robot model} The setup is the same as Appendix \ref{apx:exp:env:robot}. The only difference is that we now have a workspace limit for each finger.  

\subsubsection{Task mechanics} 
We use quasi-static models (as described in Appendix \ref{apx:exp:env:mechanics}) or force closure \cite{liu1999forceclosure}. 

\subsubsection{Feasibility Check} includes workspace limit check for fingertips, task mechanics check, and finger relocation check. 

Features, action probability, reward, value estimation, and search parameters are the same as the Manipulation with Environment Interactions task.

\subsection{Robot Experiment: Dexterous DDHand}

\subsubsection{Dexterous DDHand Overview}
Dexterous DDHand is a direct-drive hand with 4 Dofs. It has two fingers and each finger has 2 Dofs for planar translation motions. Each fingertip is a horizontal rod. As a result, we use two endpoints of the rod to approximate the line contact. 
We provide the planner with the forward and inverse kinematics of the hand. We also provide a contact relocation planner, which follows the object surface (5mm above the object surface) and goes to the new contact location. 

\subsubsection{Feasibility Checks} include inverse kinematics check, collision check, finger relocation force check, finger relocation path check (are there collisions on the relocation path), and task mechanics check.

Task mechanics, features, action probability, reward, value estimation, and search parameters are the same as the Manipulation with Extrinsic Dexterity task.

\subsubsection{Execution}

Given a planned fingertip trajectory, we compute the robot joint trajectory using inverse kinematics and execute it with robot joint position control. 
In order to ensure some contact force, we shift the end-effector trajectory in the environment contact normal direction for 
\begin{equation}
    \label{eqn:apx:stiffness}
    \Delta \mathrm{position} = \frac{\text{Desired contact force}}{\mathrm{Stiffness}}
\end{equation}
where the stiffness can be tuned due to the direct-drive property.

The execution was conducted in an open-loop manner, meaning that there was no object pose estimation or force control involved.
The system was calibrated to ensure that the initial object pose errors are kept within a tolerance of 1 mm. We chose not to provide a formal success rate in our report since this number lacks significance due to its dependency on the accuracy of our manual calibration process. However, as a point of reference, with an initial pose precision of 1 mm, we estimate a success rate of approximately 4 out of 5 attempts.

\subsection{Robot Experiment: Delta Array}

\subsubsection{Delta Array System Overview}
The array of soft delta robots is a research platform for the development of multi-robot cooperative dexterous manipulation skills. The system is comprised of 64 soft linear delta robots arranged in an 8x8 hexagonal tessellating grid. Each 3D printed soft delta linkage is actuated using 3 linear actuators to give 3 degrees of translational freedom with a workspace of 3.5cm radius in the X, and Y axes and 10cm in Z-axis. The links are compliant with high elasticity and low hysteresis, with a soft 3D printed fingertip-like end-effector attached to it. We simplify the workspace of each delta robot to be a cylinder with a 2.5cm radius and 6cm height. 
%To keep the plans within the scope of planar manipulation and reduce modeling errors. 

We provide the forward and inverse kinematic models to the planner. While running the planner, the IK check is simplified to a workspace limit check (if the contact point is in the cylinder workspace). We only perform collision checks for the fingertips, not the links. While doing the actual execution of the plans, we use inverse kinematics to calculate the robot joint trajectory from the contact point trajectory. In order to ensure some contact force, we shift the end-effector trajectory in the same way as \autoref{eqn:apx:stiffness}, where the stiffness is manually calibrated.

%Since the planner plans in simulation, it can generate contacts with switching locations across the shape of the object in simulation. But to allow for these contact mode switches in real world, 
%We develop a contact relocation method that generates switching motions of the delta robots by computing normal vectors to the center of mass of the object and aligning with the new position of the contact to allow for the next mode trajectory rollout. 
We relocate contacts by letting the delta robot to leave the contact in the contact normal direction, go around the edge of the workspace, and come to the new contact in its normal direction. 
The entire plan is executed in open-loop. Although delta robots may not offer a high level of accuracy and repeatability, their passive compliance allows for minor deviations from the planned trajectory to be accommodated.

\subsubsection{Feasibility Check} include workspace limit check, collision check, task mechanics check.

Task mechanics, features, action probability, reward, value estimation, and search parameters are the same as the Manipulation with Extrinsic Dexterity task.

\vspace{1cm}
\section{RRT for rolllout}
\label{apx:rrt}

The RRT process is summarized in Algorithm \ref{alg:rrt}. 
The inputs are the current object pose $\boldsymbol{x_{\text{current}}}$, selected contact mode $\boldsymbol{m_{\text{selected}}}$, and the object goal pose $\boldsymbol{x_{\text{current}}}$. If it can find a solution, it outputs a trajectory from $\boldsymbol{x_{\text{current}}}$ to $\boldsymbol{x_{\text{current}}}$. Every point on the trajectory is $(x, m)$, where $x \in \mathrm{SE(3)}$ is an object pose, $\boldsymbol{m}$ is an environment contact mode. 

At each iteration, \textsc{sample-random-object-pose} sample a new object pose $\boldsymbol{x_{\text{extend}}} \in \mathrm{SE(3)}$. 
We find the nearest neighbor $\boldsymbol{x_{\text{near}}}$ of $\boldsymbol{x_{\text{current}}}$, and attempt to extend it towards $\boldsymbol{x_{\text{current}}}$ (line 5 - 15, Algorithm \ref{alg:rrt}). Each extension is performed under the guidance of a contact mode. If $\boldsymbol{x_{\text{near}}}$ happens to be $\boldsymbol{x_{\text{current}}}$, we let the contact mode be $\boldsymbol{m_{\text{selected}}}$ chosen by Level 1 MCTS. Otherwise, the function \textsc{select-contact-mode} will select the contact mode(s) to perform the extension under. The procedure \textsc{extend-with-contact-mode} extends $\boldsymbol{x_{\text{near}}}$ towards $\boldsymbol{x_{\text{current}}}$ under the guidance of a selected contact mode $\boldsymbol{m}$ through projected forward integration.

\begin{algorithm}[h]
\footnotesize
\caption{RRT for Expansion and Rollout}\label{alg:rrt}
\begin{algorithmic}[1]
\Procedure{RRT-explore}{$\boldsymbol{x_{\text{current}}}$, $\boldsymbol{m_{\text{selected}}}$, $\boldsymbol{x_{\text{current}}}$}
\While{resources left and the goal is not reached}
    \State $\boldsymbol{x_{\text{rand}}} \gets$ \Call{sample-random-object-pose}{$\boldsymbol{x_{\text{goal}}}, \boldsymbol{p_{\text{sample}}}$}
    \State $\boldsymbol{x_{\text{near}}} \gets$ \Call{nearest-neighbor}{$\boldsymbol{x_{\text{rand}}}$}
    \If{$\boldsymbol{x_{\text{near}}} = \boldsymbol{x_{\text{current}}}$}
    \State $\mathcal{M} \gets \{\boldsymbol{m_{\text{selected}}}\}$
    \Else
    \State $\mathcal{M} \gets$ \Call{select-contact-modes}{$\boldsymbol{x_{\text{near}}}, \boldsymbol{x_{\text{rand}}}$}
    \EndIf
    \For {$\boldsymbol{m} \in \mathcal{M}$}
    \State $\boldsymbol{x_{\text{new}}} \gets$ \Call{extend-with-contact-mode}{$\boldsymbol{x_{\text{near}}}, \boldsymbol{x_{\text{rand}}}, \boldsymbol{m}$}
    \If {$\boldsymbol{x_{\text{new}}} \neq \text{null}$}
    \State \Call{add-to-rrt-tree}{$\boldsymbol{x_{\text{new}}}$, $\mathcal{T}_{\text{rrt}}$}
    \EndIf
    \EndFor 
\EndWhile
\State solution-path $\gets$ \Call{backtrack}{$\boldsymbol{x_{\text{goal}}}, \mathcal{T}_{\text{rrt}}$}
\State \textbf{return} solution-path
\EndProcedure

\Procedure{select-contact-mode}{$\boldsymbol{x_{\text{near}}}, \boldsymbol{x_{\text{rand}}}$}
\State $\boldsymbol{p_{env}} \gets$ \Call{environment-contact-point-detection}{$\boldsymbol{x_{\text{near}}}$}
\State $\mathcal{M}_{env} \gets$ \Call{enumerate-contact-modes}{$p_{env}\boldsymbol{p_{env}} $}
\For{all $\boldsymbol{m} \in \mathcal{M}_{env}$}
    \If{\Call{extend-feasibility-check}{m, $\boldsymbol{x_{\text{near}}}, \boldsymbol{x_{\text{rand}}}$}}
        \State $\mathcal{M} \gets \{M, \boldsymbol{m}\}$
    \EndIf
\EndFor
\State \textbf{return} $\mathcal{M}$

\EndProcedure
\Procedure{extend-with-contact-mode}{$\boldsymbol{x_{\text{near}}}, \boldsymbol{x_{\text{rand}}}, \boldsymbol{m}$}
\State $\boldsymbol{x_{\text{now}}} = \boldsymbol{x_{\text{near}}}$
\While{true}
\If{not \Call{extend-feasibility-check}{m, $\boldsymbol{x_{\text{now}}}, \boldsymbol{x_{\text{rand}}}$}}
    \State break
\EndIf
\State $\boldsymbol{v} \gets $ \Call{velocity-under-mode}{m, $\boldsymbol{x_{\text{now}}}, \boldsymbol{x_{\text{rand}}}$}
\If {$\boldsymbol{v}$ close to zero}
    \State break
\EndIf
\State \note{Projected forward integration}
\State $\boldsymbol{x_{\text{now}}} \gets $ \Call{integrate}{$\boldsymbol{x_{\text{now}}}, \boldsymbol{v}$}
\State $\boldsymbol{x_{\text{now}}} \gets$ \Call{project-to-contacts-maintained}{$\boldsymbol{x_{\text{now}}}, \boldsymbol{m}$}
\If {encounter new contacts}
\State break
\EndIf
\EndWhile
\State \textbf{return} $\boldsymbol{x_{\text{now}}}$
\EndProcedure

\end{algorithmic}
\end{algorithm}

Next, we explain all the functions in detail. 

\textsc{sample-random-object-pose} sample a new object pose $\boldsymbol{x_{\text{extend}}} \in \mathrm{SE(3)}$. The probability of $\boldsymbol{x_{\text{current}}}$ being the goal pose is $\boldsymbol{p_{\text{sample}}}$, while the probability of it being a random object pose in $\mathrm{SE(3)}$ is $1 - \boldsymbol{p_{\text{sample}}}$. We can specify the range limit for the random sample of the object pose. 

\textsc{nearest-neighbor} finds the closest object pose to $\boldsymbol{x_{\text{current}}}$ in the tree. The distance between two object poses is computed as $w_t * d_t + w_r * d_r$. $w_t$ and $w_r$ are the weights for translation and rotation. $d_t$ is the Euclidean distance between their locations, and $d_r$ is the angle difference between two rotations. A simple way to compute $d_r$ is to first compute the rotation between two poses $R_{\text{diff}} = R_1R_2^T$, and convert $R_{\text{diff}}$ to axis-angle representation and let $d_r$ be equal to the angle. In general, the users need to adjust the weights according to how important object orientation or position is important in the task. Not much tuning is needed. 
In our experiment, we scale the object sizes such that the average length of their bounding boxes is about 1 to 10. In this case, one can set the weights using this rule: normal (1), not very important (0.5), not important at all (0.1). 

\textsc{select-contact-mode} first enumerates all contacting-separating contact modes, then filters out infeasible mode through \textsc{extend-feasibility-check}, and finally returns the set of all feasible modes. 

\textsc{extend-feasibility-check} has two options in implementation. The first option involves storing the robot contacts during the search process. If the current robot contacts pass the feasibility check in Section~\ref{sec:method-level2-feasibility}, the check is deemed successful. However, if they fail, the check can still be considered successful if a sampled set of feasible robot contacts can be generated, ensuring a feasible transition from the current contacts. The current contacts are then updated accordingly. 
The second option finds if there exists a set of robot contacts that satisfy the feasibility check. Unlike the first option, this method does not retain information about robot contacts. Instead, it is consider successful if it can sample any set of robot contacts, as long as they pass the feasibility check. The second option is more relaxed as it does not take into account previous robot contacts and transitions.

\textsc{extend-with-contact-mode} extends $\boldsymbol{x_{\text{near}}}$ towards $\boldsymbol{x_{\text{current}}}$ as much as possible under constraints posted by contact mode $\boldsymbol{m}$. 
\textsc{velocity-under-mode} solves for the object velocity that get $\boldsymbol{x_{\text{near}}}$ as close as possible to $\boldsymbol{x_{\text{current}}}$ with respect to the velocity constraints introduced by $\boldsymbol{m}$. We then integrate the object pose for a small step in the direction of constrained object velocity, and project the new object pose back to the contacts that needed to be maintained. 

In \textsc{project-to-contacts-maintained}, the contact mode $\boldsymbol{m}$ needs to be maintained. We first perform contact detection on the object. We then project the object pose back to where the maintaining contacts in $\boldsymbol{m}$ have zero signed distances. 

\subsection{Complexity}

We combined a vanilla RRT algorithm with contact mode enumeration and projected forward integration. The increase in complexity is associated with the need to enumerate contact modes for each object modes. We also need to store these contact modes and extend them to generate new nodes. 

Let's say we have $N$ nodes in the RRT, where each node is associated with an object configuration in the space. 
For each node, we need to enumerate the contact modes. If there are $M$ contact points detected in the current object configuration, the number of contact modes and the time complexity of enumerating them is $O(M^d)$, where $d$ is the DoFs of free motion of the object. For an object free to move in the 3D space, $d=6$. 
Since we need to perform the \textsc{extend} process for each node and save the contact modes, this contact mode enumeration process adds space and time complexity $O(M^d)$ to each node. 

Thus, the space complexity is $O(N * (M^d))$. The RRT extend process has a time complexity of $O(M^d)$ because it needs to perform projected integration ($O(1)$) for all the contact modes. While the projected forward integration can take considerably longer with collision check and contact projection. However, these calculations do not depend on the number of vertices already in the tree. The rest stays the same as a vanilla RRT --- $O(N)$ for the time complexity of RRT nearest neighbor query; $O(1)$ for inserting new vertices. 

Note that the current bottleneck of our computation time is not the time complexity that grows with the number of nodes. It is the projected forward integration process because we need to perform collision checking and contact projection. While these calculations do not depend on the number of vertices already in the tree, they consist of almost 80 percent of the computation time.

\end{appendices}

\end{document}